\documentclass[1pt,twocolumn,letter]{article}
\usepackage[a4paper, left=0.75in, right=0.75in, top=1in, bottom=1in]{geometry}
\newsavebox\CBox
\def\textBF#1{\sbox\CBox{#1}\resizebox{\wd\CBox}{\ht\CBox}{\textbf{#1}}}
\usepackage[table,xcdraw]{xcolor}
\usepackage{graphicx}
\usepackage{booktabs}
\usepackage{multirow}
\usepackage[ruled,vlined]{algorithm2e}

\usepackage{algorithm}

\usepackage{wrapfig} 
\usepackage{amsmath,amssymb} 
\usepackage{mathtools} 
\usepackage{enumitem} 
\usepackage{makecell} 
\usepackage[colorlinks=true,linkcolor=red,citecolor=green,urlcolor=blue]{hyperref}

\usepackage{paralist}
\usepackage{amsmath,amssymb}
\usepackage{tabularx}
\usepackage{comment}
\usepackage{caption}
\usepackage{cleveref}
\usepackage{pdfpages}
\usepackage{times}

\definecolor{myblue}{rgb}{0.21,0.49,0.74} 
\definecolor{lightgreen}{rgb}{0.6, 0.9, 0.6}
\definecolor{lightyellow}{rgb}{1.0, 1.0, 0.6}
\definecolor{lightorange}{rgb}{1.0, 0.8, 0.6}
\definecolor{lightpurple}{rgb}{0.8, 0.7, 1.0}
\definecolor{lightpink}{rgb}{1.0, 0.8, 0.9}
\definecolor{lightgray}{rgb}{0.9, 0.9, 0.9}
\definecolor{darkred}{rgb}{0.65, 0.0, 0.0} 
\definecolor{darkblue}{rgb}{0, 0.4, 0.75} 
\usepackage{authblk}

\title{Incrementally Learning Multiple Diverse Data Domains\\ via Multi-Source Dynamic Expansion Model}

\author[1]{Runqing Wu}
\author[2]{Fei Ye\thanks{Corresponding author. Email: \texttt{feiye@uestc.edu.cn}}}
\author[2]{Qihe Liu}
\author[3]{Guoxi Huang}
\author[2]{Jinyu Guo}
\author[2]{, Rongyao Hu}

\affil[1]{School of Mechanical Engineering, Huazhong University of Science and Technology, China}
\affil[2]{School of Information and Software Engineering, University of Electronic Science and Technology, China}
\affil[3]{University of Bristol, England}

\date{}

\begin{document}

\maketitle

\begin{abstract}
Continual Learning seeks to develop a model capable of incrementally assimilating new information while retaining prior knowledge. However, current research predominantly addresses a straightforward learning context, wherein all data samples originate from a singular data domain. This paper shifts focus to a more complex and realistic learning environment, characterized by data samples sourced from multiple distinct domains. We tackle this intricate learning challenge by introducing a novel methodology, termed the Multi-Source Dynamic Expansion Model (MSDEM), which leverages various pre-trained models as backbones and progressively establishes new experts based on them to adapt to emerging tasks. Additionally, we propose an innovative dynamic expandable attention mechanism designed to selectively harness knowledge from multiple backbones, thereby accelerating the new task learning. Moreover, we introduce a dynamic graph weight router that strategically reuses all previously acquired parameters and representations for new task learning, maximizing the positive knowledge transfer effect, which further improves generalization performance. We conduct a comprehensive series of experiments, and the empirical findings indicate that our proposed approach achieves state-of-the-art performance.

\noindent \textbf{Details:} The implementation of our proposed framework is available at \href{https://github.com/LexRider/MSDEM}{https://github.com/LexRider/MSDEM}.

\end{abstract}

\section{Introduction}
Continual Learning (CL), often referred to as lifelong learning, seeks to develop a model that can consistently acquire new concepts while retaining previously learned information \cite{LifeLong_review}. Nevertheless, contemporary deep learning models frequently experience considerable performance decline in the context of continual learning, primarily due to catastrophic forgetting \cite{LifeLong_review}. This issue arises because these models lack the necessary mechanisms to safeguard against information loss when adapting to new tasks. Given its advantageous characteristics, continual learning holds significant practical applications across various fields, including autonomous driving, robotic navigation, and medical diagnostics.

Continual learning research has led to the development of various technologies aimed at addressing the issue of network forgetting. These can be categorized into three main approaches: first, rehearsal-based methods, which focus on optimizing a compact memory buffer to retain numerous essential examples \cite{TinyLifelong,RainbowMemory}; second, dynamic expansion frameworks that facilitate the automatic construction and integration of new hidden layers and nodes into an existing backbone to capture new information \cite{Adanet,ompactingPicking}; and third, regularization-based methods that incorporate an additional regularization term into the primary objective function to mitigate significant changes to many previous and crucial network parameters \cite{MeasureForgetting,OptimizingNeural}.

Among these technologies, utilizing a memory buffer to retain numerous critical examples stands out as the most prevalent approach in continual learning; however, it struggles to accommodate an increasing number of tasks. Conversely, dynamic expansion models excel in this demanding continual learning paradigm by adaptively generating a new sub-model to tackle each new task \cite{Adanet,ompactingPicking}. Furthermore, recent research has suggested leveraging a pre-trained Vision Transformer (ViT) \cite{Vit} as a foundational backbone, allowing for the rapid construction of a new expert with minimal parameters to swiftly adapt to new tasks \cite{BoostingCL}. Nonetheless, these methodologies typically concentrate on a singular data domain and depend on a single pre-trained backbone \cite{BoostingCL}, which limits their ability to generalize effectively across a sequence of diverse data domains where both the domain and class may shift unpredictably.

In this paper, we introduce an innovative dynamic expansion framework, referred to as the Multi-Source Dynamic Expansion Model (MSDEM), designed to tackle the challenges of class and domain shifts in multi-domain continual learning. The core concept of our proposed methodology is to integrate knowledge retained by various backbones trained on distinct data sources into a cohesive optimization framework, with the objective of delivering robust generalization representations for the experts. Specifically, we present a novel Dynamic Expandable Attention Mechanism (DEAM) that regulates representations from multiple backbones through an attention mechanism. In contrast to existing attention-based approaches \cite{Vit}, which extract relevant information from the pixel space, our proposed attention mechanism can dynamically assess the significance and contribution of each backbone during the learning of new tasks, thereby effectively exploring prior knowledge to promote the new task learning.

Furthermore, numerous tasks and domains often share analogous semantic information, making it essential to leverage previously acquired knowledge to facilitate future task learning. To achieve this aim, we propose a novel Dynamic Graph Weight Router (DGWR) strategy, which oversees and optimizes a graph relation matrix to regulate all previously learned information during the learning of new tasks. The DGWR approach effectively reuses critical past parameters and representations, significantly enhancing the learning of new tasks and resulting in improved generalization performance.

We run a series of experiments utilizing various intricate datasets, and the empirical findings indicate that the proposed methodology attains state-of-the-art performance in more demanding continual learning scenarios while utilizing fewer parameters. Our contributions can be categorized into four parts~: \begin{inparaenum}[(1)]
\item This paper introduces a novel MSDEM framework to deal with a sequence of diverse data domains by exploring knowledge from several backbones trained on different data sources;
\item We propose a novel dynamic expandable attention mechanism to selectively transfer knowledge from several backbones, which maximizes the transfer learning effects;
\item We propose a novel DGWR approach to effectively reuse all previously learned parameters and representations to promote future task learning, leading to an improved generalization performance;
\item We construct a more realistic and challenging continual learning experiment and the empirical results demonstrate that the proposed approach achieves the state-of-the-art performance.
\end{inparaenum}

\section{Related Work}
\textBF{Rehearsal-based techniques} represent a fundamental and widely adopted strategy to mitigate network forgetting in continuous learning, as highlighted in the recent literature \cite{CLBlurry}.  This approach focuses on retaining a substantial number of essential past instances and reintroducing them during the learning of new tasks \cite{CLBlurry,Co2l,NoSelection,CLMutual,LearnAdd,LifeLong_combination,OnlineStructuredLaplace,GCR,DualPrototypeCL}.  Consequently, the selection of samples is pivotal in ensuring optimal performance of rehearsal-based methods. Additionally, the integration of a memory buffer system can be effectively aligned with regularization-based techniques, with the objective of further enhancing the model's efficacy \cite{plasticityCL,OptimizingNeural,TinyLifelong,GradientEpisodic,KernelCL,CLBit,CLNullSpace,VCL,Uncertainty_CL,FlattingCL,NPCL}. Another approach to implement the memory system is to train a deep generative model such as Variational Autoencoders (VAEs) \cite{VAE} or Generative Adversarial Networks (GANs) \cite{GAN} for preserving and producing past examples to relieve network forgetting \cite{Lifelong_VAE,GenerativeLifelong,Generative_replay,LifelongGAN,Sddgr}. These methods can avoid the data privacy issues caused by the memory buffer system.

\vspace{2pt}
\noindent
\textBF{Knowledge distillation techniques} focus on transferring the knowledge preserved by a static teacher module to a small-sized student module \cite{KD_Review,Distilling_nets}. The Knowledge Distillation (KD) approach can also be applied to continual learning for addressing network forgetting. Specifically, the KD approach in continual learning treats the previously and currently learned model as a teacher and a student module, respectively. Minimizing the distance between the teacher and student outputs can relieve network forgetting \cite{Lwf}. Furthermore, the KD methodology can be integrated with rehearsal-based methods into a cohesive optimization framework, which can further enhance model performance, as demonstrated in \cite{icarl}, namely Incremental Classifier and Representation Learning (iCaRL). Specifically, iCaRL employs a novel nearest-mean-of-exemplars classification approach that bolsters the classifier's resilience to variations in data representations. In addition, another studies introduce a new self-KD technology, aiming to preserve previously acquired features and representations, which can address network forgetting \cite{Co2l}.

\vspace{2pt}
\noindent
\textBF{Dynamic network architecture.} Although rehearsal and knowledge distillation (KD) technologies have achieved significant performance in continual learning, they can only perform well on a small number of tasks and can not address the more complex learning environment. Recent studies have developed a dynamic expandable framework to deal with a long sequence of tasks. Specifically, this framework automatically creates and adds new sub-models and hidden layers into a unified backbone when learning a new task, in which all previously learned parameters are frozen to preserve all prior knowledge \cite{Adanet,ompactingPicking,LearnAdd,ProgressiveNN,BatchEnsemble,OnlineLearning,ForgetFree,EffecientFeature}. As a result, the dynamic expandable framework can maintain good performance on all previous tasks without forgetting and are able to learn a new task effectively through a dynamic expansion process \cite{ProgressiveNN}. Furthermore, the recent popular backbone, called Vision Transformers (ViT) \cite{Vit}, has also been explored as a sub-network into a dynamic expansion framework, which can achieve better performance than the CNN based dynamic expansion framework \cite{MetaVitCL,Dytox}. We provide additional information for the related work section in \textBF{Appendix-A} from Supplementary Material (SM).

\section{Methodology}
\subsection{Problem Statement}

In continual learning, a model is assumed to be trained in a dynamically changed learning environment. Specifically, the model can only access the training samples from the current task learning while all previous tasks are unavailable. Let $D^s_i = \{ {\bf x}^i_j, {\bf y}^i_j \}^{n^i}_{j=1}$ and $D^t_i = \{ {\bf x}^{t,i}_j, {\bf y}^{t,i}_j \}^{n^{t,i}}_{j=1}$ be the $i$-th training and testing dataset, respectively, where $n^i$ and $n^{t,i}$ denote the total number of samples in the training set $D^s_i$ and the testing set $D^t_i$, respectively. ${\bf x}^{t,i}_j \in {\mathcal{X}}$ and ${\bf y}^{t,i}_j \in {\mathcal{Y}}$ denote the $j$-th testing sample and its corresponding class label, respectively. ${\mathcal{X}} \in {\mathbb R}^{d_{x}}$ and ${\mathcal{Y}} \in {\mathbb R}^{d_{y}}$ are the data and label space with the dimension $d_x$ and $d_y$, respectively. In a class-incremental learning paradigm, a training dataset ${D}^s_i$ is usually divided into $C_i$ parts $\{ {D}^s_i(1), \cdots,{D}^s_i(C_i) \}$, where each subset ${D}^s_i(j)$ contains data samples from a single or several adjacent categories. Let $\{T_1,\cdots,T_{C_i}\}$ be a set of tasks, where each task ${T_j}$ is associated with the training dataset ${D}^s_i(j)$. At a certain task learning (${T}_j$), we can only access the samples from ${D}^s_i(j)$ while all previous datasets $\{ 
{D}^s_i(1) ,\cdots, {D}^s_i(j-1) \}$ are unavailable. 
Most existing continual learning studies only consider to incrementally learn new categories within a single data domain. However, in a more realistic learning environment, new data samples can be drawn from entirely different data domains. Let $\{ D^s_1,\cdots,D^s_t \}$ be a set of $t$ different datasets/domains, where each dataset $D^s_i$ can be divided into $C_i$ parts. A data stream $S$ can be formed using the following process~:
\begin{equation}
    \begin{aligned}
        S = \{ D^s_1(1),\cdots,D^s_1(C_1),\cdots,D^s_t(C_t) \}\,.
    \end{aligned}
\end{equation}
Learning the data stream $S$ remains a considerable challenge since it involves shifts in both the class and domain. When the total number of tasks is finished, we evaluate the model's performance on all testing datasets. 

\begin{figure*}[t]
  \centering
  \includegraphics[page=1, width=\textwidth, trim=60 175 60 80, clip]{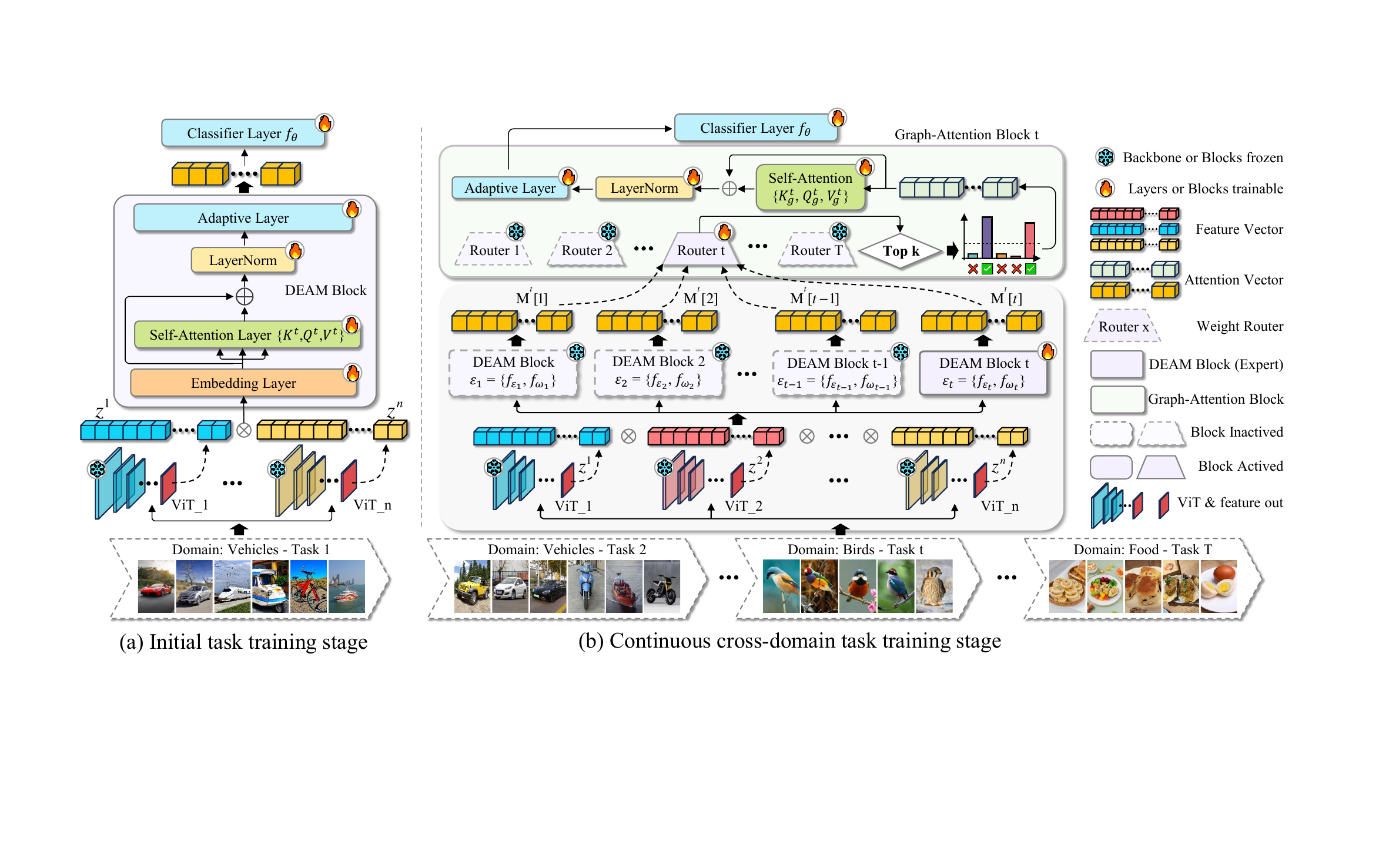}
    \captionsetup{skip=4pt} 
  \caption{Overall framework of the proposed method. (a) During the initial training task, task \(T_1\) is used as an input sample to multiple backbones, generating individual feature outputs that are concatenated to form a fused feature vector. This fused vector is then processed through a multi-head attention module for feature integration, followed by a classifier head to produce the final result. (b) In the subsequent training tasks across multiple domains, the attention modules from previous tasks are retained as experts and frozen. The output feature vectors from all experts are fused and then fed into a router, where weight allocation and Top-k selection are applied to identify the most important experts for knowledge integration. The resulting fused vector is then processed through graph attention for the final prediction.}
  \label{fig:pdf-multi-column}
  \vspace{-15pt}
\end{figure*}

\subsection{Framework Overview}

Existing research typically proposes the introduction of a new independent expert within a mixture system or the utilization of a single pre-trained Vision Transformer (ViT) as a foundational backbone to initialize an expert with minimal parameters for learning a new task. However, many of these approaches focus solely on a single pre-trained backbone that encompasses semantically rich knowledge from one or a few data domains, which limits their applicability to unknown and entirely distinct data domains. In this paper, we introduce an innovative dynamic expansion framework that incorporates multiple pre-trained ViT backbones trained on samples from diverse sources, referred to as the Multi-Source Dynamic Expansion Model (MSDEM). This model demonstrates robust generalization capabilities across various data domains. We present a comprehensive overview of the network architecture for the proposed framework in \cref{fig:onecol}, which comprises several network modules, detailed in the following.

\vspace{2pt}
\noindent
\textBF{The multi-source backbones.} Utilizing multiple backbones that are trained on diverse datasets and data distributions can yield semantically rich and robust representations, thereby enhancing the model's generalization capabilities in continual learning. Let $\{f_{\theta_1}, \cdots,f_{\theta_{t'}} \}$ represent a collection of $t'$ distinct backbones, each trained on varying data domains and datasets. Each backbone $f_{\theta_j} \colon \mathcal{X} \to \mathcal{Z}$ is constructed using the pre-trained Vision Transformer (ViT) \cite{Vit}, which takes an image ${\bf x} \in {\mathcal{X}}$ as input and produces a feature vector ${\bf z} \in {\mathcal{Z}}$, where ${\mathcal{Z}} \in {\mathbb R}^{d_z}$ denotes the feature space with dimension $d_z$, and $\theta_j$ signifies the parameter set of the $j$-th backbone. Given the substantial output dimension of each pre-trained ViT backbone, we utilize only the class token to minimize feature dimensions and computational expenses. For any input ${\bf x}$, we can leverage all pre-trained backbones to extract a potent representation by~:
\begin{equation}
    \begin{aligned}
    {\bf z}^f = {\bf z}^1 \otimes {\bf z}^2 \otimes,\cdots, \otimes \, {\bf z}^{t'}\,, 
    \label{combinedFeatures}
    \end{aligned}
\end{equation}
\noindent where ${\bf z}^j$ is given by the $j$-th backbone $f_{\theta_j}$ and $\otimes$ denotes to combine two feature vectors into a single one. ${\bf z}^f$ is an augmented feature vector over the feature space ${\mathcal{Z}}^f \in {\mathbb R}^{d_z \times t'}$. 

\vspace{2pt}
\noindent
\textBF{The expert module.} While the pre-trained backbone is capable of delivering robust representations, it cannot directly leverage these features for predictive tasks. In this study, we introduce a method to dynamically construct and integrate a new expert module within the proposed dynamic expansion framework to address the challenges of learning new tasks. Specifically, for a designated new task ${T}_j$, we develop a new expert module ${\mathcal{E}}_j$, which comprises an adaptive module $f_{\xi_j} \colon \mathcal{Z}^f \to \mathcal{Z}^e$ designed to acquire a task-specific representation, alongside a linear classifier $f_{\omega_j} \colon \mathcal{Z}^e \to \mathcal{Y}$ intended to discern a decision-making pattern. The adaptive module $f_{\xi_j}$ of the $j$-th expert ${\mathcal{E}}_j$ takes an augmented feature ${\bf z}^f$ as input and produces a feature vector ${\bar{\bf z}}^j$ within the feature space ${\mathcal{Z}}^f \in \mathbb{R}^{d_e}$, where $d_e$ denotes the dimensionality of the features. The prediction process for a given input ${\bf x}$ utilizing the $j$-th expert is articulated as follows~:
\begin{equation}
    \begin{aligned}
    y' =
\arg\max(
    {\rm Softmax}(
         {\bf W}^{\rm T}_{\omega_j} {\bar{\bf z}}^j ))\,,
         \label{classifierEq}
    \end{aligned}
\end{equation}
\noindent where ${\bf W}_{\omega_j}$ denotes the weight matrix of the classifier $f_{\omega_j}$ and ${\rm Softmax}(\cdot)$ represents a Softmax function. ${\bf W}^{\rm T}_{\omega_j}$ represents the matrix transpose. 

\subsection{Dynamic Expandable Attention Mechanism}
In the process of acquiring the knowledge from a new task, certain pre-trained backbones may encompass semantically relevant representations that facilitate the learning of the new task, thereby enhancing their contribution to this learning process. Merely aggregating representations from various backbones, as outlined in Eq.~\eqref{combinedFeatures}, fails to effectively leverage prior knowledge for the new task acquisition. To tackle this challenge, we introduce an innovative dynamic expandable attention mechanism that autonomously assesses the significance of each pre-trained ViT backbone. Specifically, our proposed method can automatically generate and incorporate a new attention module upon the creation of a new expert, enabling the development of an expert-specific attention behaviour.

We assume that the proposed framework has already learnt $(t-1)$ experts $\{ {\mathcal{E}}_1,\cdots,{\mathcal{E}}_{t-1} \}$. When learning a new task (${T}_{t}$), we dynamically create three trainable weight matrices ${\bf K}^{t}$, ${\bf Q}^{t}$ and ${\bf V}^{t}$, to regulate all previously learned representations during the new task learning. For a given input ${\bf x}$, we can get a combined representation ${\bf z}^f$ using Eq.~\eqref{combinedFeatures} and employ the attention mechanism to process ${\bf z}^f$~:

\begin{equation}
    \begin{aligned}
&{\hat{\bf Q}}^{t} = {\bf Q}^{t} {\bf z}^f , {\hat{\bf K}}^{t} = {\bf K}^{t} {\bf z}^f,\\&  {\hat{\bf V}}^{t} = {\bf V}^{t} {\bf z}^f\,.
\label{attentionOperator}
    \end{aligned}
\end{equation}

The resulting weight matrices ${\hat{\bf Q}}^{t}$, ${\hat{\bf K}}^{t}$ and ${\hat{\bf V}}^{t}$ are used to calculate the attention map by~:

\begin{equation}
    \begin{aligned}
{\bf z}^{t}_{\rm att} = {\rm Softmax}( {\hat{\bf Q}}^{t} ( {\hat{\bf K}}^{t} / \sqrt{d_k} )) {\hat{\bf V}}^{t} \,,
    \end{aligned}
\end{equation}
\noindent where ${\bf z}^{t}_{\rm att}$ is an attention map, which is used as the input of the adaptive module $f_{\xi_{t}}$ of the new expert (${\mathcal{E}}_{t}$). Additionally, we only update the weight matrices $\{{\bf K}^{t}, {\bf Q}^{t}, {\bf V}^{t} \}$ during the new task learning (${T}_{t}$) while all previous weight matrices $\{ {\bf K}^{1}, {\bf Q}^{1}, {\bf V}^{1},\cdots,{\bf K}^{t-1}, {\bf Q}^{t-1}, {\bf V}^{t-1}  \}$ are frozen to avoid forgetting all previous attention behavrious.

\subsection{Dynamic Graph Weight Router}
\begin{figure}[t]
  \centering
\includegraphics[page=2, width=\columnwidth, trim=200 220 200 210, clip]{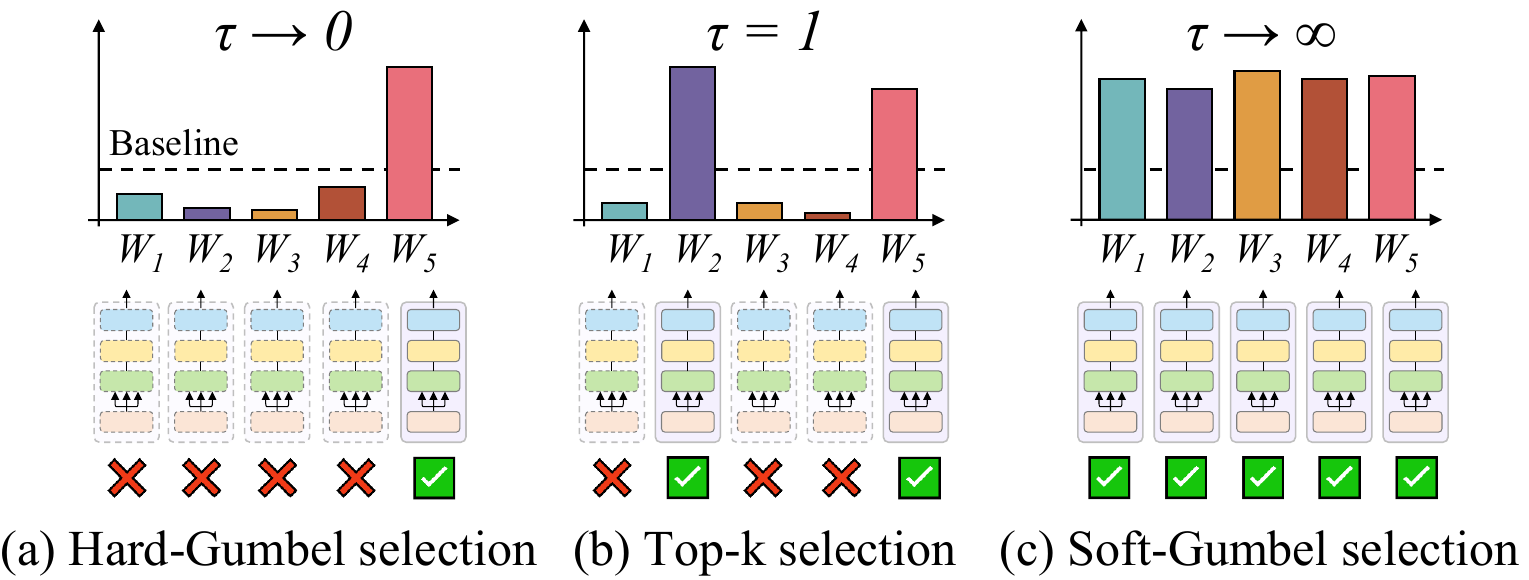}
  \captionsetup{skip=4pt} 
  \caption{The expert selection process with different values of \(\tau\). (a) When the temperature is low, the selection approaches a one-hot vector, selecting only the expert for the current task. (b) When the temperature is set close to 1, it performs Top-k selection, where \(k\) is learned during training rather than manually constrained. (c) When the temperature is high, all experts are selected.}
  \label{fig:onecol}
  \vspace{-10pt}
\end{figure}

Most current studies in continual learning primarily concentrate on examining the active model parameters to acquire new tasks, which limits their ability to capture comprehensive statistical information. In this paper, we propose leveraging numerous essential previously acquired network parameters and representations to bolster the learning capacity for future tasks. Furthermore, given that each expert assimilates knowledge from distinct data domains, employing all previously learned network parameters may not be advantageous for new task acquisition. We tackle this challenge by introducing an innovative dynamic adaptive weight generation method that dynamically constructs and develops weight routers to selectively identify several key experts for new task learning. We assume that the proposed dynamic expansion framework has already established $(t-1)$ experts $\{ {\mathcal{E}}_1,\cdots,{\mathcal{E}}_{t-1} \}$ during the $(t-1)$-th task learning phase. We conceptualize each expert as a node within a graph structure and present a graph relation matrix ${\bf C} \in {\mathbb{R}}^{(t-1,t-1)}$ to characterize the interrelations among experts, where ${\bf C}(i,j)$ signifies the relationship from the $j$-th expert to the $i$-th expert. Upon encountering a new task ($T_{t}$), we first establish a new expert module ${\mathcal{E}}_{t}$ and extend the relation matrix to ${\bf C} \in {\mathbb{R}}^{(t,t)}$. Subsequently, we extract the relation vector ${\bf M}^{t} = \{ {\bf M}^{t}[1],\cdots, {\bf M}^{t}[t] \}$ from ${\bf C}(t)$, which represents all elements of the $t$-th row of ${\bf C}$, corresponding to the expert (${\mathcal{E}}_t$). To effectively select experts for new task learning, we propose utilizing the Gumbel-Softmax distribution to generate the weight router, expressed as~:
\begin{equation}
    \begin{aligned}
    {\widehat{\bf M}}^{t}[k] = \frac{\exp\left((\log({{\bf M}}^{t}[k] + {\epsilon}_n) + {\epsilon}_u) / \tau \right)}{\sum\nolimits_{j=1}^{t} \exp\left((\log({{\bf M}}^{t}[j] + {\epsilon}_n) + {\epsilon}_u) / \tau \right)}\,,
\label{eq:gumble}
    \end{aligned}
\end{equation}
\noindent
where ${\epsilon}_n \sim \mathcal{N}(0, \sigma^2 I)$ is sampled from a normal noise distribution to enhance the robustness of weight optimization while encouraging the model to explore different expert combinations more extensively during the early stages of training. ${\epsilon}_u = -\log(-\log(U))$ and $U \sim \text{Uniform}(0, 1)$. $\tau$ is a temperature parameter, controlling the smoothness of the sampling distribution. We describe the expert selection in Fig.~\ref{fig:onecol}. Compared to a rigid Top-k selection scheme, adjusting the temperature parameter of the Gumbel-Softmax allows for tuning between hard and soft selection. In practice, the number of selected experts is determined by the results of the optimization process. 
By using Eq.~\eqref{eq:gumble}, we can get the selection weights $\{ {\widehat{\bf M}}^{t}[1],\cdots, {\widehat{\bf M}}^{t}[t] \}$, which can be used to regulate the representations extracted by all previously learned experts.


Then we combine all normalized representations and the feature extracted from the adaptive module $f_{\xi_{t}}$ into a compact representation, expressed as~:
\begin{equation}
    \begin{aligned}
    {\bf Z}^{t} = \sum\nolimits^t_{j=1} \left(  {\bar{\bf z}}^j  {\widehat{\bf M}}^{t}[j]  \right)\,.
    \label{finalRepresentation}
    \end{aligned}
\end{equation}
We employ the attention mechanism to further regulate the representation ${\bf Z}^t$, resulting in~:
\begin{equation}
    \begin{aligned}
&{\hat{\bf Q}}^{t}_g = {\bf Q}^{t}_g {\bf Z}^t , {\hat{\bf K}}^{t}_g = {\bf K}^{t}_g {\bf Z}^t,\\&  {\hat{\bf V}}^{t}_g = {\bf V}^{t}_g {\bf Z}^t\,,
\label{GraphOperator}
    \end{aligned}
\end{equation}
\noindent where ${\bf Q}^{t}_g$, ${\bf K}^{t}_g$, and ${\bf V}^{t}_g$ are the weight matrices of the attention mechanism. We can get the attention results by~:
\begin{equation}
    \begin{aligned}
{\bf Z}^{t}_{\rm a} = {\rm Softmax}( {\hat{\bf Q}}^{t}_g ( {\hat{\bf K}}^{t}_g / \sqrt{d^{'}_{\rm total}} )) {\hat{\bf V}}^{t}_g \,,
\label{GetGraphOutput}
    \end{aligned}
\end{equation}
\noindent where ${d^{'}_{\rm total}}$ refers to the dimension of ${{\bf Z}^t}$. The prediction process for the $t$-th expert is reformulated as~:
\begin{equation}
    \begin{aligned}
    y' =
\arg\max(
    {\rm Softmax}(
         {\bf W}^{\rm T}_{\omega_{t}} {\bf Z}^{t}_{\rm a} ))\,,
        \label{classifierEq2}
    \end{aligned}
\end{equation}
\noindent where ${\bf W}_{\omega_{t}}$ denotes weight matrix of the classifier $f_{\omega_t}$. Only the parts ${\bf C}(t)$ of the relation matrix ${\bf C}$ is optimized while other parts are frozen during the new task learning, which can avoid forgetting the previously learned router.

\subsection{Algorithm Implementation}
In this section, we provide the learning procedure of the proposed framework in \cref{fig:pdf-multi-column} while the 
pseudocode is provided in \textBF{Algorithm 1}, which can be summarized into three stages, described in the following.

\vspace{2pt}
\noindent
\textBF{Step 1 (The construction process).} When learning a new task ${T}_t$, we dynamically create a new expert module $ {\mathcal{E}}_t = \{ f_{\xi_t}, f_{\omega_t} \} $ based on the pre-trained backbones and ${\bf K}^t_g, {\bf Q  }^t_g, {\bf V}^t_g$ of graph block.

\vspace{2pt}
\noindent
\textBF{Step 2 (The dynamic expandable attention mechanism).} We first create the attention parameters $\{ {\bf K}^{t},{\bf Q}^{t},{\bf V}^{t} \}$, which can be used to regulate the augmented feature ${\bf z}^f$ using Eq.~\eqref{attentionOperator}, resulting in $\{ {\hat{\bf K}}^{t},{\hat{\bf Q}}^{t},{\hat{\bf V}}^{t}  \}$. 

\vspace{2pt}
\noindent
\textBF{Step 3 (The dynamic graph weight router).} We expand the relation matrix ${\bf C}$ and create the router $\{{\widehat{\bf M}}^{t}[1],\cdots,{\widehat{\bf M}}^{t}[t] \}$ for the task ${T_t}$ using Eq.~\eqref{eq:gumble}. Then, we can obtain the final representation $ {\bf Z}^{t}_{\rm a}$ using Eq.~\eqref{GetGraphOutput}. 
 
\vspace{2pt}
\noindent
\textBF{Step 4 (The parameter update).} To optimize the proposed framework, we introduce to employ the cross-entropy loss, defined as~:
\begin{equation}
    \begin{aligned}
{\mathcal{L}}_{\rm CE} &=  
    \sum\nolimits_{c=1}^{K} {\bf y}[{c}] 
 \log \big\{ {\rm Softmax}(
         {\bf W}^{\rm T}_{\omega_{t}} {\bf Z}^{t}_{\rm a} )[c] \big\} \,,
         \label{crossEntropy}
    \end{aligned}
\end{equation}
\noindent where $K$ is the total number of categories and ${\bf y}[c]$ is the $c$-th dimension of the class label. ${\rm Softmax}(
         {\bf W}^{\rm T}_{\omega_{t}} {\bf Z}^{t}_{\rm a} )[c]$ denotes the $c$-th dimension of the probability vector. During the current task learning ($T_t$), we only update the parameter sets $\{ \xi_t, \omega_t, {\bf K}^t_g, {\bf Q  }^t_g, {\bf V}^t_g, {\bf K}^t, {\bf Q  }^t, {\bf V}^t, {\bf M}^{t} \}$ of the current expert ${\mathcal{E}}_t$ using Eq.~\eqref{crossEntropy}. 
\vspace{-5pt}

\begin{table*}[t]
\centering
\caption{Performance comparison of MSDEM and SOTA models in a dual-domain task configuration. "Average" denotes mean performance across all tasks, while "Last" shows the performance on the final task. Except for StarPrompt comparisons (trained for 3 epochs), all models are trained for 1 epoch and compared with non-StarPrompt SOTA models. MSDEM$^2$ and MSDEM$^3$ represent configurations with 2 and 3 ViT backbones, respectively, using 32-head multi-head attention. All results are averaged over 10 independent runs.}
\vspace{-8pt}
\fontsize{8.0pt}{11pt}\selectfont 
\setlength{\tabcolsep}{0.5pt} 
\begin{tabularx}{\textwidth}{l*{12}{>{\centering\arraybackslash}X}} 
\toprule
\textbf{Method} & \multicolumn{2}{c}{\textbf{TinyImage-Birds}} & \multicolumn{2}{c}{\textbf{Birds-TinyImage}} & \multicolumn{2}{c}{\textbf{Cifar10-Birds}} & \multicolumn{2}{c}{\textbf{Birds-Cifar10}} & \multicolumn{2}{c}{\textbf{Cifar100-Birds}} & \multicolumn{2}{c}{\textbf{Birds-Cifar100}} \\
 & Average & Last & Average & Last & Average & Last & Average & Last & Average & Last & Average & Last \\
\midrule

DER \cite{CL_DarkKD} & 5.5\text{\scriptsize ±0.48} & 95.4\text{\scriptsize ±0.72} & 26.3\text{\scriptsize ±0.64} & 92.5\text{\scriptsize ±0.93} & 5.5\text{\scriptsize ±0.25} & 96.0\text{\scriptsize ±0.42} & 55.9\text{\scriptsize ±0.34} & 96.7\text{\scriptsize ±0.81} & 10.0\text{\scriptsize ±0.99} & 77.4\text{\scriptsize ±0.55} & 22.5\text{\scriptsize ±0.88} & 75.1\text{\scriptsize ±0.67} \\
DER++ \cite{CL_DarkKD} & 93.3\text{\scriptsize ±0.81} & 95.8\text{\scriptsize ±0.93} & 93.3\text{\scriptsize ±0.62} & 94.6\text{\scriptsize ±0.89} & 94.2\text{\scriptsize ±0.45} & 96.4\text{\scriptsize ±0.75} & 98.9\text{\scriptsize ±0.82} & 96.4\text{\scriptsize ±0.32} & 89.9\text{\scriptsize ±0.67} & 80.4\text{\scriptsize ±0.95} & 96.2\text{\scriptsize ±0.79} & 95.6\text{\scriptsize ±0.56} \\
DER+++refresh \cite{der+++refresh}& 93.3\text{\scriptsize ±0.92} & 95.2\text{\scriptsize ±0.89} & 93.4\text{\scriptsize ±0.25} & 93.5\text{\scriptsize ±0.63} & 94.8\text{\scriptsize ±0.88} & 96.4\text{\scriptsize ±0.24} & 98.8\text{\scriptsize ±0.91} & 96.4\text{\scriptsize ±0.79} & 90.6\text{\scriptsize ±0.55} & 78.8\text{\scriptsize ±0.43} & 96.0\text{\scriptsize ±0.73} & 94.8\text{\scriptsize ±0.34} \\
MoE-2E/1R \cite{BoostingCL}& 24.5\text{\scriptsize ±0.85} & 91.2\text{\scriptsize ±0.31} & 28.7\text{\scriptsize ±0.63} & 91.1\text{\scriptsize ±1.21} & 19.7\text{\scriptsize ±0.52} & 99.1\text{\scriptsize ±0.54} & 33.9\text{\scriptsize ±0.43} & 96.8\text{\scriptsize ±0.35} & 33.4\text{\scriptsize ±0.52} & 96.2\text{\scriptsize ±0.32} & 37.4\text{\scriptsize ±0.22} & 93.5\text{\scriptsize ±0.82} \\
MoE-22E/10R \cite{BoostingCL}& 26.3\text{\scriptsize ±0.94} & 88.9\text{\scriptsize ±0.47} & 25.0\text{\scriptsize ±0.71} & 94.6\text{\scriptsize ±1.31} & 23.3\text{\scriptsize ±0.68} & 96.8\text{\scriptsize ±0.44} & 33.6\text{\scriptsize ±0.98} & 97.4\text{\scriptsize ±1.53} & 31.5\text{\scriptsize ±0.91} & 95.8\text{\scriptsize ±0.43} & 36.6\text{\scriptsize ±0.61} & 94.5\text{\scriptsize ±1.36} \\
StarPrompt-1$^{st}$ \cite{starprompt}& 96.8\text{\scriptsize ±0.42} & 96.1\text{\scriptsize ±0.52} & 96.8\text{\scriptsize ±0.64} & 96.0\text{\scriptsize ±0.83} & 99.1\text{\scriptsize ±0.65} & 99.0\text{\scriptsize ±0.95} & 99.0\text{\scriptsize ±0.73} & 97.1\text{\scriptsize ±0.87} & 97.1\text{\scriptsize ±0.92} & \textbf{99.7\text{\scriptsize ±0.82}} & 97.0\text{\scriptsize ±0.45} & 95.4\text{\scriptsize ±0.79}\\
StarPrompt-2$^{nd}$ \cite{starprompt}& 97.5\text{\scriptsize ±0.77} & 97.3\text{\scriptsize ±0.51} & 97.7\text{\scriptsize ±0.65} & \textbf{96.4\text{\scriptsize ±0.43}} & 97.0\text{\scriptsize ±0.23} & \textbf{99.7\text{\scriptsize ±0.95}} & 93.8\text{\scriptsize ±0.76} & 97.3\text{\scriptsize ±0.92} & 98.0\text{\scriptsize ±0.43} & \textbf{99.7\text{\scriptsize ±0.56}} & 98.1\text{\scriptsize ±0.95} & 97.3\text{\scriptsize ±0.63} \\
\rowcolor{lightgray}
StarPrompt \cite{starprompt}& \textbf{97.8\text{\scriptsize ±0.48}} & \textbf{98.9\text{\scriptsize ±0.82}} & \textbf{97.8\text{\scriptsize ±0.79}} & 96.3\text{\scriptsize ±0.91} & \textbf{99.2\text{\scriptsize ±0.37}} & 99.0\text{\scriptsize ±0.71} & \textbf{99.2\text{\scriptsize ±0.46}} & 98.0\text{\scriptsize ±1.01} & \textbf{98.3\text{\scriptsize ±0.59}} & 96.8\text{\scriptsize ±1.32} & \textbf{98.2\text{\scriptsize ±0.52}} & 97.6\text{\scriptsize ±0.92} \\
RanPac \cite{RanPAC} & 93.8\text{\scriptsize ±0.88} & 91.2\text{\scriptsize ±0.31} & 94.1\text{\scriptsize ±0.65} & 95.9\text{\scriptsize ±0.43} & 98.9\text{\scriptsize ±0.74} & 93.1\text{\scriptsize ±0.53} & 98.7\text{\scriptsize ±0.92} & 98.7\text{\scriptsize ±0.42} & 95.4\text{\scriptsize ±0.95} & 88.6\text{\scriptsize ±0.32} & 95.4\text{\scriptsize ±0.32} & \textbf{98.6\text{\scriptsize ±0.32}}\\
Dap \cite{dap}& 92.9\text{\scriptsize ±0.72} & 95.0\text{\scriptsize ±0.89} & 92.4\text{\scriptsize ±0.52} & 93.4\text{\scriptsize ±0.41} & 83.4\text{\scriptsize ±0.67} & 97.9\text{\scriptsize ±0.88} & 90.7\text{\scriptsize ±0.42} & \textbf{99.0\text{\scriptsize ±0.32}} & 90.4\text{\scriptsize ±0.52} & 94.8\text{\scriptsize ±0.42} & 90.6\text{\scriptsize ±0.68} & 98.0\text{\scriptsize ±0.72} \\
\midrule
\rowcolor{lightpink}
MSDEM$^2$(Ours) & \textbf{97.8\text{\scriptsize ±0.23}} & \textbf{99.0\text{\scriptsize ±0.78}} & \textbf{97.7\text{\scriptsize ±0.92}} & 95.0\text{\scriptsize ±0.63} & \textbf{99.3\text{\scriptsize ±0.65}} & 97.9\text{\scriptsize ±0.79} & \textbf{99.4\text{\scriptsize ±0.21}} & 97.5\text{\scriptsize ±0.54} & \textbf{98.0\text{\scriptsize ±0.52}} & \textbf{99.3\text{\scriptsize ±0.42}} & \textbf{98.1\text{\scriptsize ±0.37}} & \textbf{99.1\text{\scriptsize ±0.61}} \\
Rel.ER vs RanPac & $\downarrow$ 64.52\% & $\downarrow$ 88.63\% & $\downarrow$ 61.02\% & $\uparrow$ 21.95\% & $\downarrow$ 36.36\% & $\downarrow$ 69.56\% & $\downarrow$ 53.84\% & $\uparrow$ 92.31\% & $\downarrow$ 56.52\% & $\downarrow$ 93.86\% & $\downarrow$ 58.69\% & $\downarrow$ 35.71\% \\
MSDEM$^3$(Ours) & 96.7\text{\scriptsize ±0.74} & 94.9\text{\scriptsize ±0.67} & 96.7\text{\scriptsize ±0.42} & \textbf{99.6\text{\scriptsize ±0.41}} & 99.1\text{\scriptsize ±0.83} & 97.5\text{\scriptsize ±0.56} & 99.1\text{\scriptsize ±0.56} & \textbf{99.0\text{\scriptsize ±1.36}} & 97.1\text{\scriptsize ±0.83} & 99.1\text{\scriptsize ±0.55} & 97.0\text{\scriptsize ±1.11} & 94.0\text{\scriptsize ±0.76} \\
Rel.ER vs RanPac & $\downarrow$ 46.77\% & $\downarrow$ 42.04\% & $\downarrow$ 44.07\% & $\downarrow$ 90.24\% & $\downarrow$ 18.18\% & $\downarrow$ 63.77\% & $\downarrow$ 30.77\% & $\downarrow$ 23.08\% & $\downarrow$ 36.95\% & $\downarrow$ 92.11\% & $\downarrow$ 34.78\% & $\uparrow$ 328.5\% \\
\midrule
\rowcolor{lightpink}
MSDEM$^2$(Ours)-3ep & \textbf{98.1\text{\scriptsize ±0.52}} & \textbf{99.4\text{\scriptsize ±0.32}} & \textbf{98.0\text{\scriptsize ±0.85}} & 96.3\text{\scriptsize ±0.67} & \textbf{99.6\text{\scriptsize ±0.41}} & \textbf{99.8\text{\scriptsize ±0.63}} & \textbf{99.7\text{\scriptsize ±0.31}} & 97.9\text{\scriptsize ±0.55} & \textbf{98.3\text{\scriptsize ±0.29}} & 99.3\text{\scriptsize ±0.19} & \textbf{98.4\text{\scriptsize ±0.54}} & 97.3\text{\scriptsize ±0.34} \\
Rel.ER vs StarPrompt & $\downarrow$ 13.64\% & $\downarrow$ 45.45\% & $\downarrow$ 9.09\% & $\downarrow$ 0.00\% & $\downarrow$ 63.64\% & $\downarrow$ 79.88\% & $\downarrow$ 62.50\% & $\uparrow$ 5.33\% & $\downarrow$ 0.00\% & $\downarrow$ 78.13\% & $\downarrow$ 11.11\% & $\uparrow$ 12.55\% \\

\bottomrule
\end{tabularx}
\label{tab:shortdomainAcc}
\vspace{-15pt}
\end{table*}

\begin{algorithm}[t]
\caption{The training of the proposed framework.}
 \textBF{Input:} The total number of tasks $N$; The model's parameters; \;
 
\textBF{Output:} The model's parameters \;

\For{$t < N$}{

    \textBF{Step 1 (The construction process).}\;

    Build a new expert $ {\mathcal{E}}_t = \{ f_{\xi_t}, f_{\omega_t} \} $
             \;

    \textBF{Step 2 (The dynamic expandable attention mechanism).}\;

        Build the attention parameters $\{ {\bf K}^{t},{\bf Q}^{t},{\bf V}^{t} \}$ \;
        
        ${\hat{\bf Q}}^{t} = {\bf Q}^{t} {\bf z}^f, {\hat{\bf K}}^{t} = {{\bf K}}^{t}  {\bf z}^f, {\hat{\bf V}}^{t} = {\bf V}^{t} {\bf z}^f $ \; 

        ${\bf z}^{j}_{\rm att} = {\rm Softmax}( {\hat{\bf Q}}^{t} ( {\hat{\bf K}}^{t} / \sqrt{d_k} )) {\hat{\bf V}}^{t}$ \;

    \textBF{Step 3 (The dynamic graph weight router).}\;

Build the router $\{{\widehat{\bf M}}^{t}[1],\cdots,{\widehat{\bf M}}^{t}[t] \}$  \;

$\{  {\Tilde{\bf z}}^c \,|\, 
{\Tilde{\bf z}}^c = {\bar{\bf z}}^c {\widehat{\bf M}}^{t}[j] \,, j = 1,\cdots,t   \}$ \;

${\bf Z}^{t} = {\sum\nolimits^{t}_{j=1}}\{ {\Tilde{\bf z}}^{t}  \} $ ;

Get ${\bf Z}^{t}_{\rm a}$ using Eq.~\eqref{GetGraphOutput} \;

    \For{$t < n'$}{
        \textBF{Step 4 (The parameter update).}\;

        Get a new data batch ${\bf X}_t$ from the current task \;
                
        Optimize
        $\{ \xi_t, \omega_t, {\bf M}^{t}  \}$ by Eq.~\eqref{crossEntropy} \;
    
 Optimize  $\{ {\bf K}^t, {\bf Q  }^t, {\bf V}^t, {\bf K}^t_g, {\bf Q  }^t_g, {\bf V}^t_g   
 \}$ by Eq.~\eqref{crossEntropy}  \;
 } 
}

\end{algorithm}

\section{Experiments}
\vspace{-4pt}
\subsection{Experimental setting}
\vspace{-2pt}
\noindent
\textbf{Datasets:} We evaluate the model performance in a continual learning setting across multiple domains, using the TinyImageNet \cite{TinyImageNet}, CIFAR-100 \cite{CIFAR10}, CIFAR-10 \cite{CIFAR10}, and Birds 525 Species datasets. We provide additional experiment settings in \textBF{Appendix-B} from SM.

\noindent
\textbf{Metrics:} To assess and compare the model performance across multi-domain settings, we use two metrics: "Average" and "Last." The "Average" metric measures the model’s mean accuracy across tasks in a specific scenario, while "Last" indicates the accuracy on the final task. We calculate the average accuracy on all testing samples.

\noindent
\textbf{Implementation:} We use three different ViT models as backbones: ViT-B/16 pretrained on ImageNet-21K, ViT-B/16 pretrained on ImageNet-21K and fine-tuned on ImageNet-1K, and pretrained ViT-L/14. All three models are frozen during training and inference to retain domain-specific prior knowledge. 
For the classifier, router, and all attention layers, we consider to employ three different Adam optimizers with tailored learning rates and scheduler parameters, aiming to achieve optimal performance.

\begin{table*}[ht]
\centering
\caption{Performance comparison of MSDEM and SOTA in 3-domain and 4-domain configurations, summarizing average performance across all tasks and performance on the final task.}
\vspace{-8pt}
\fontsize{8.5pt}{11pt}\selectfont 
\setlength{\tabcolsep}{10pt} 
\begin{tabularx}{\textwidth}{l*{8}{>{\centering\arraybackslash}X}} 
\toprule
\textbf{Method} & \multicolumn{2}{c}{\textbf{Tiny-Cifar10-Birds}} & \multicolumn{2}{c}{\textbf{Tiny-Cifar100-Birds}} & \multicolumn{2}{c}{\textbf{Tiny-C100-Birds-C10}} & \multicolumn{2}{c}{\textbf{Average}} \\
 & Average & Last & Average & Last & Average & Last & Average & Last \\
\midrule
DER \cite{CL_DarkKD}& 6.23\text{\scriptsize ±0.34} & 95.0\text{\scriptsize ±0.56} & 9.46\text{\scriptsize ±0.45} & 94.9\text{\scriptsize ±0.42} & 14.92\text{\scriptsize ±0.56} & 95.01\text{\scriptsize ±0.66} & 17.36 & 90.89\\
DER++ \cite{CL_DarkKD}& 92.04\text{\scriptsize ±0.44} & 95.91\text{\scriptsize ±0.39} & 87.47\text{\scriptsize ±0.53} & 94.93\text{\scriptsize ±0.37} & 88.82\text{\scriptsize ±0.37} & \textbf{99.51\text{\scriptsize ±0.20}} & 92.68 & 94.39\\
DER+++refresh \cite{der+++refresh}& 94.82\text{\scriptsize ±0.50} & 96.35\text{\scriptsize ±0.52} & 90.56\text{\scriptsize ±0.40} & 78.83\text{\scriptsize ±0.46} & 91.19\text{\scriptsize ±0.54} & 94.43\text{\scriptsize ±0.38} & 93.71 & 91.64\\
MoE-2E+1R \cite{BoostingCL}& 31.22\text{\scriptsize ±0.36} & 92.00\text{\scriptsize ±0.41} & 28.83\text{\scriptsize ±0.43} & 91.36\text{\scriptsize ±0.40} & 27.55\text{\scriptsize ±0.51} & 92.33\text{\scriptsize ±0.42} & 29.47 & 93.73\\
MoE-22E+10R \cite{BoostingCL}& 34.55\text{\scriptsize ±0.63} & 94.56\text{\scriptsize ±1.17} & 31.22\text{\scriptsize ±0.36} & 31.22\text{\scriptsize ±0.36} & 95.11\text{\scriptsize ±0.78} & 30.01\text{\scriptsize ±0.33} & 37.46 & 80.42\\
StarPrompt-1$^{st}$ \cite{starprompt}& 96.71\text{\scriptsize ±0.54} & \textbf{99.15\text{\scriptsize ±0.63}} & 96.03\text{\scriptsize ±0.33} & \textbf{99.02\text{\scriptsize ±0.16}} & 95.97\text{\scriptsize ±0.46} & 97.14\text{\scriptsize ±0.71} & 97.17 & 97.62\\
StarPrompt-2$^{nd}$ \cite{starprompt}& 89.45\text{\scriptsize ±0.53} & 92.04\text{\scriptsize ±0.47} & 84.36\text{\scriptsize ±0.82} & 95.06\text{\scriptsize ±0.39} & 85.49\text{\scriptsize ±0.20} & 93.77\text{\scriptsize ±0.71} & 93.49 & 96.50\\
\rowcolor{lightgray}
StarPrompt \cite{starprompt}& \textbf{97.70\text{\scriptsize ±0.65}} & 99.12\text{\scriptsize ±0.77} & \textbf{97.01\text{\scriptsize ±0.15}} & 98.01\text{\scriptsize ±1.00} & \textbf{97.39\text{\scriptsize ±0.35}} & 97.76\text{\scriptsize ±0.71} & \textbf{98.06} & \textbf{98.14} \\
RanPac \cite{RanPAC}& 93.92\text{\scriptsize ±0.48} & 91.10\text{\scriptsize ±0.35} & 94.15\text{\scriptsize ±0.38} & 92.32\text{\scriptsize ±0.76} & 94.14\text{\scriptsize ±0.39} & 95.15\text{\scriptsize ±0.60} & 95.38 & 93.85\\
Dap \cite{dap}& 94.48\text{\scriptsize ±0.51} & 92.65\text{\scriptsize ±0.45} & 92.77\text{\scriptsize ±0.39} & 95.51\text{\scriptsize ±0.40} & 91.62\text{\scriptsize ±0.47} & 95.83\text{\scriptsize ±0.42} & 91.03 & 96.49\\
\midrule
\rowcolor{lightpink}
MSDEM$^2$\text{(Ours)} & \textbf{97.74\text{\scriptsize ±0.46}} & \textbf{98.55\text{\scriptsize ±0.49}} & \textbf{96.92\text{\scriptsize ±0.54}} & \textbf{99.23\text{\scriptsize ±0.53}} & \textbf{97.09\text{\scriptsize ±0.22}} & \textbf{97.95\text{\scriptsize ±0.36}} & \textbf{98.00} & \textbf{98.48}\\
Rel.ER vs RanPac & $\downarrow$ 62.8\% & $\downarrow$ 83.7\% & $\downarrow$ 47.3\% & $\downarrow$ 89.9\% & $\downarrow$ 50.3\% & $\downarrow$ 57.7\% & $\downarrow$ 56.7\% & $\downarrow$ 50.6\% \\
MSDEM$^3$\text{(Ours)} & 97.15\text{\scriptsize ±0.38} & 95.7\text{\scriptsize ±0.44} & 96.15\text{\scriptsize ±0.41} & 95.55\text{\scriptsize ±0.50} & 95.92\text{\scriptsize ±0.34} & 97.33\text{\scriptsize ±1.38} & 97.21 & 96.96\\
Rel.ER vs RanPac & $\downarrow$ 53.1\% & $\downarrow$ 51.7\% & $\downarrow$ 34.2\% & $\downarrow$ 42.1\% & $\downarrow$ 81.6\% & $\downarrow$ 44.9\% & $\downarrow$ 39.6\% & $\downarrow$ 50.6\%  \\
\midrule
\rowcolor{lightpink}
MSDEM$^2$(Ours)-3ep & \textbf{98.15\text{\scriptsize ±0.14}} & \textbf{99.12\text{\scriptsize ±0.83}} & \textbf{97.59\text{\scriptsize ±0.36}} & \textbf{98.38\text{\scriptsize ±1.11}} & \textbf{97.65\text{\scriptsize ±0.66}} & \textbf{97.97\text{\scriptsize ±0.91}} & \textbf{98.39} & \textbf{98.38} \\
Rel. ER vs StarPrompt & $\downarrow$ 19.6\% & $\downarrow$ 0.00\% & $\downarrow$ 14.5\% & $\downarrow$ 18.6\% & $\downarrow$ 9.96\% & $\downarrow$ 9.37\% & $\downarrow$ 17.0\% & $\downarrow$ 12.9\%  \\

\bottomrule
\end{tabularx}
\label{tab:longdomainAcc}
\vspace{-5pt}
\end{table*}

\subsection{Comparison with SOTA}
\textbf{SOTA Categorization.} In this section, we present a comparative analysis of our method against various SOTA approaches in continual learning. Specifically, we consider the experience replay-based methods such as DER \cite{CL_DarkKD}, along with its enhanced variants DER++ \cite{CL_DarkKD} and DER+++refresh \cite{der+++refresh}. We also consider to compare our approach with the dynamic expansion model such as the MoE adapter-based models utilizing a mixture of experts framework where we distinguish between a lower-bound configuration (MoE-2E/1R, comprising 2 experts and 1 router) and an upper-bound configuration (MoE-22E/10R, with 22 experts and 10 routers) \cite{BoostingCL}. Additionally, we also consider employing the prompt-based learning models as the baseline such as the StarPrompt \cite{starprompt}, which maintains a balance between new and previous tasks through prompt injection and generated replay. 
Finally, we compare methods based on another type of continual learning methods, including Random Packing (RanPac) \cite{RanPAC}, which employs a random grouping mechanism, and Data Augmentation Prompt (Dap) \cite{dap}, which incorporates data augmentation to enhance the retention of prior knowledge in continual learning scenarios. All models employing memory replay strategies (DER, DER++, DER+++refresh) use the same backbone. To maintain consistency in our experimental setup, we configure them with a dual-ViT model, unfreezing the last two blocks of each ViT to provide a fine-tuning parameter space. The memory replay buffer size is uniformly set to the Maximum 5120.

\noindent
\textbf{Multi-domain Task Incremental Learning.} In this experiment setting, we consider employing six two-domain scenarios, two three-domain scenarios, and one four-domain scenario, with evaluations based on two performance metrics: “Average” and “Last”. Specifically, in the two-domain scenarios, we explore various domain order combinations to assess the generalization performance of various models under different domain configurations. Additionally, we examine both dual-ViT and triple-ViT model strategies to evaluate whether more pre-trained backbones can improve the model's generalization performance.  It is noteworthy that the prompt-based StarPrompt, along with its generative replay mechanism, reuses the training samples from the current task 2–4 times during training to strengthen resistance to forgetting. Therefore, we also include the results achieved by the proposed approach using 3 training epochs for a fair comparison.

\noindent
\textBF{The results.} We present the classification results of our method with other SOTA approaches in \cref{tab:shortdomainAcc} and \cref{tab:longdomainAcc}. The empirical results clearly illustrate that our method (denoted as "Ours") achieves superior average performance under the dual ViT model setup across nearly all task configurations. Notably, memory replay-based methods such as DER and mixture-of-experts models like MoE exhibit relatively poor performance in these multi-domain task scenarios. Although they achieve relatively high “Last” scores, indicating strong performance on the current task, they show limited resistance to forgetting.

In the dual-domain configuration, our method shows a $16.98\%$ improvement in the Average metric and $15.95\%$ in the Last metric over StarPrompt, the baseline performance ceiling. In the three-domain and four-domain configurations, the improvements are 15.06\% and 6.2\%, respectively. All comparisons are based on training for 3 epochs, where our model achieves great performance gains as the number of domains increases. Even with training limited to a single epoch, Our method consistently outperforms all 
SOTAs except StarPrompt on the Average metric across all task configurations, with results closely matching StarPrompt-1$^{st}$ and StarPrompt-$2^{nd}$ in the Cifar100-Birds and Birds-Cifar100 tasks, respectively.

\begin{figure}[t]
  \centering
  \includegraphics[page=3, width=\columnwidth, trim=170 80 170 80, clip]{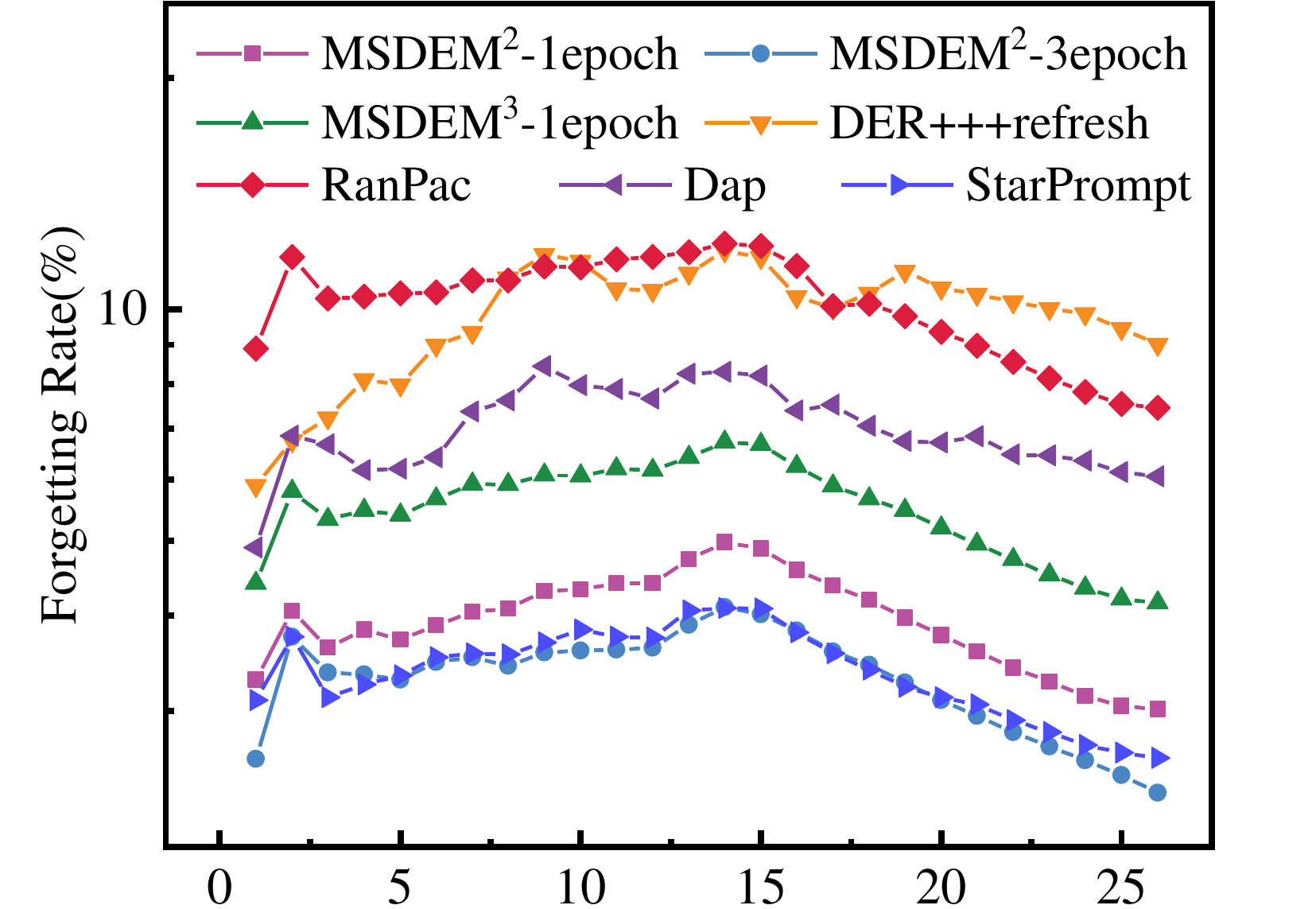}
  \captionsetup{skip=4pt} 
  \caption{In the T-C100-B-C10 task configuration, we compare the forgetting curves of MSDEM against SOTA methods. An additional 3-epoch MSDEM variant is included for comparison with StarPrompt. Our model consistently outperforms all SOTA approaches across all tasks.}
  \vspace{-15pt}\label{fig:sotaforgettingline}
\end{figure}

When considering to use a single training epoch, our method outperforms RanPac by $67.49\%$ in the average metric and $70.44\%$ in the last metric in the dual-domain configuration. In the three-domain and four-domain scenarios, these improvements are $62.28\%$ and $78.88\%$, respectively. It is also worth noting that, although the triple ViT configuration is outperformed by the dual ViT setup, the proposed approach still maintains a leading edge over other SOTA methods across nearly all task configurations. This finding underscores the effectiveness of our model while suggesting that increasing the number of ViTs does not necessarily lead to continuous performance gains. Our forgetting lines with SOTA are summarized in Fig.~\ref{fig:sotaforgettingline}. 

\begin{table*}[t]
\centering
\caption{Comparison of our method with other SOTA methods in terms of training parameters, GPU usage, CPU usage, and training time. Arrows (↑, ↓) indicate whether higher or lower values are preferred for each metric. Performance results are evaluated based on two model strategies: dual ViT and triple ViT backbones. All results are obtained in the "Tiny-Birds" task scenario on the same hardware environment (RTX 4090 with 24GB memory) and represent the average of five runs on the same task.}
\vspace{-8pt}
\fontsize{8.2pt}{9.5pt}\selectfont 
\setlength{\tabcolsep}{3pt} 
\renewcommand{\arraystretch}{1.1} 
\begin{tabularx}{\textwidth}{l*{9}{>{\centering\arraybackslash}X}} 
\toprule
\textbf{Method} & \textbf{Train Param} $\downarrow$ & \textbf{GPU Max} $\downarrow$ & \textbf{GPU Avg} $\downarrow$ & \textbf{CPU Max} $\downarrow$ & \textbf{CPU Avg} $\downarrow$ & \textbf{Iteration $\uparrow$} & \textbf{Task Time $\downarrow$} \\
\midrule
DER++ & 42.27M & 3490 MiB & 3490 MiB & 25415 MiB & 24209 MiB & 3.22 it/s & 110.5s \\
DER+++refresh & 42.27M & 9914 MiB & 9914 MiB & 19668 MiB & 18847 MiB & 2.27 it/s & 357.74s \\
MoE-22E+10R & 64.05M & 23560 MiB & 21362 MiB & 18662 MiB & 18289 MiB & 1.93 it/s & 266.65s \\
StarPrompt-1$^{st}$ & 0.31M & 12578 MiB & 3144 MiB & 18725 MiB & 18178 MiB & 1.58 it/s & 226.19s \\
StarPrompt-2$^{nd}$ & \cellcolor{lightgray}86.11M & \cellcolor{lightgray}11610 MiB & \cellcolor{lightgray}11010 MiB & \cellcolor{lightgray}18348 MiB & \cellcolor{lightgray}17968 MiB & \cellcolor{lightgray}1.18 it/s & \cellcolor{lightgray}386.87s \\
StarPrompt & 86.41M & 10566 MiB & 10112 MiB & 9255 MiB & 8605 MiB & 2.49 it/s & 424.19s \\
RanPac & 1.49M & 3804 MiB & 3566 MiB & 18321 MiB & 17902 MiB & 3.44 it/s & 250.82s \\
Dap & 0.68M & 4420 MiB & 4420 MiB & 17764 MiB & 17224 MiB & 2.33 it/s & 147.08s \\
\midrule
MSDEM$^2$\text{(Ours)} & \cellcolor{lightpink}25.52M & \cellcolor{lightpink}1736 MiB & \cellcolor{lightpink}1560 MiB & \cellcolor{lightpink}19357 MiB & \cellcolor{lightpink}18584 MiB & \cellcolor{lightpink}8.94 it/s & \cellcolor{lightpink}41.22s \\
vs StarPrompt-2$^{nd}$ & \textcolor{darkblue}{\textbf{-70.36\% $\downarrow$}} & \textcolor{darkblue}{\textbf{-85.05\% $\downarrow$}} & \textcolor{darkblue}{\textbf{-85.83\% $\downarrow$}} &  \textcolor{darkred}{\textbf{+5.49\% $\uparrow$}} & \textcolor{darkred}{\textbf{+3.43\% $\uparrow$}} & \textcolor{darkblue}{\textbf{+657.63\% $\uparrow$}} & \textcolor{darkblue}{\textbf{-89.35\% $\downarrow$}} \\
MSDEM$^3$\text{(Ours)} & \cellcolor{lightpink}153.58M & \cellcolor{lightpink}3846 MiB & \cellcolor{lightpink}3109 MiB & \cellcolor{lightpink}21167 MiB & \cellcolor{lightpink}20538 MiB & \cellcolor{lightpink}6.67 it/s & \cellcolor{lightpink}67.70s \\
vs StarPrompt-2$^{nd}$ & \textbf{+78.35\% $\uparrow$} & \textbf{-66.87\% $\downarrow$} & \textbf{-69.25\% $\downarrow$} & \textbf{+15.36\% $\uparrow$} & \textbf{+14.30\% $\uparrow$} & \textbf{+465.25\% $\uparrow$} & \textbf{-82.50\% $\downarrow$} \\

\bottomrule
\end{tabularx}
\label{tab:performance}
\vspace{-5pt}
\end{table*}


\subsection{Ablation Study}

We provide the additional ablation results in \textBF{Appendix-C} from SM.

\noindent
\textBF{Computational Cost.} We compare the computational costs of our method with other baselines in terms of the computational costs and the number of parameters. A comparative analysis across various models is provided in \cref{tab:performance}, which reports the training parameters (M), peak and average GPU memory usage (MiB), and runtime efficiency (it/s). The proposed framework, which utilizes a dual-backbone strategy, shows a clear advantage over all current SOTA methods. Compared with a prominent SOTA method, StarPrompt-2$^{nd}$, our approach reduces training parameters by $70.36\%$, GPU memory usage by $85.83\%$, and training time by $89.35\%$. This underscores its capacity to enhance continual learning in ViT-based models while significantly alleviating computational demands during training. Although our method shows a modest $3.43\%$ increase in RAM usage compared to StarPrompt-2$^{nd}$, it remains within the typical range observed across other SOTA methods such as DER++refresh and RanPac. In addition, we evaluate the proposed approach with three ViT backbones and the empirical results show that the proposed approach requires a bit more training time while maintaining highly competitive performance. In addition, more backbones enable to expand the dimension of the embedding layer, resulting in a great increasing number of parameters. 

\begin{figure}[t]
  \centering
\includegraphics[page=4, width=\columnwidth, trim=165 170 160 180, clip]{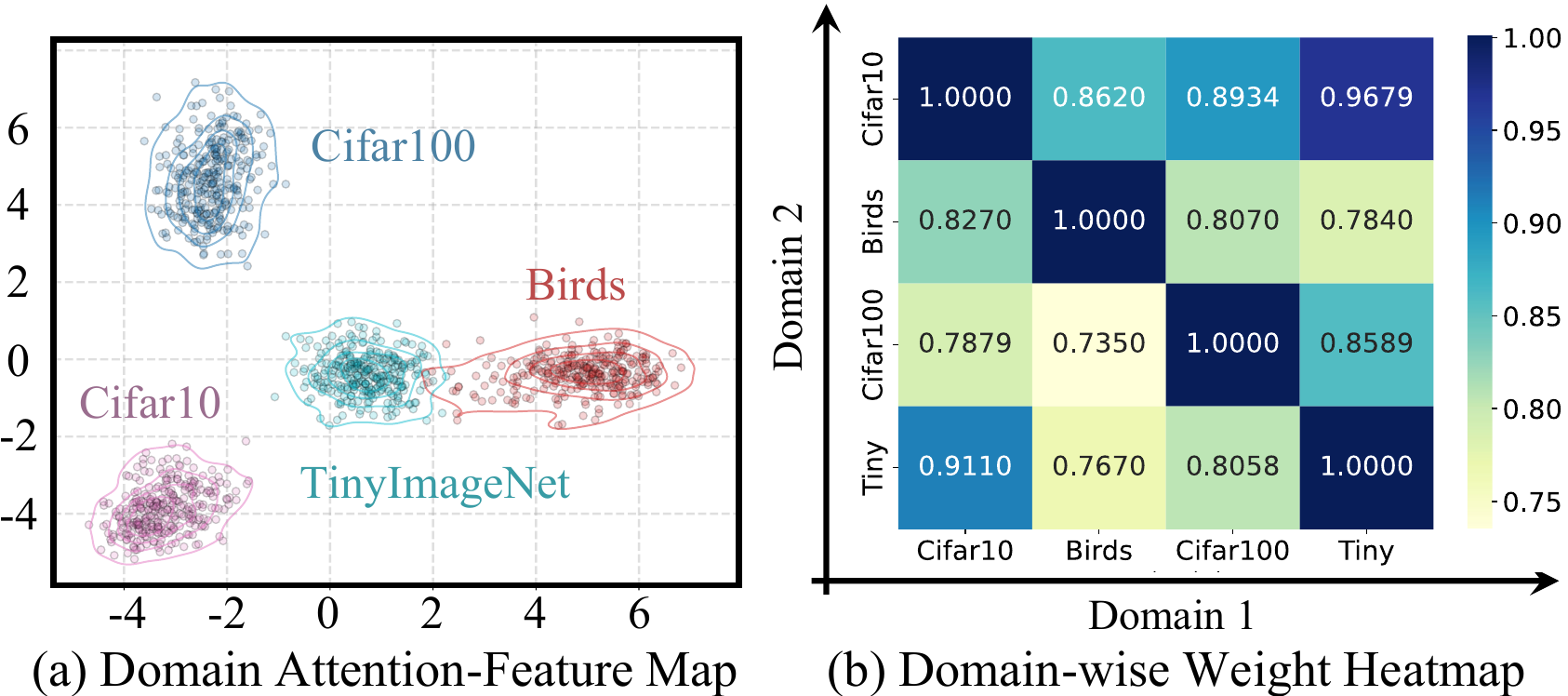}
\captionsetup{skip=4pt} 
  \caption{(a) Feature maps of trained experts across four domains, reduced via PCA. Most domains exhibit mutual independence, with the exception of the relatively close alignment between TinyImageNet and Birds; (b) Heatmap showing the knowledge weights of domain2 on domain1 across 16 permutation schemes, with each domain’s self-dependency set to 1.
  }
  \label{fig:heatmap}
  \vspace{-4pt}
\end{figure}

\noindent
\textBF{Analysis of Router.} The router mechanism is introduced to balance the extent of knowledge transfer from previous tasks during the training of the current task, as knowledge dependencies vary across domains. ~\cref{fig:heatmap}(a) presents the Feature Map derived from PCA analysis of output features from expert models trained on four domains. While these domains are generally independent, their relative distances are uneven. By permutating all domain pairs and following a training sequence from domain1 to domain2, we calculated the weight assigned to domain1’s knowledge after domain2 training, thereby quantifying domain2’s dependency on domain1. As shown in Figure ~\cref{fig:heatmap}(b), nearly all permutation schemes indicate that domain2’s dependency on domain1 shifts with their training order, suggesting that knowledge dependency between domains is asymmetric. Notably, when two domains are relatively close, such as TinyImageNet and Birds, their dependencies tend to be reciprocal.

\section{Conclusion}
\vspace{-5pt}

This paper deals with continual learning by developing a novel learning framework, which dynamically creates a new expert based on multiple backbones to learn a new task. A new dynamic expandable attention mechanism is proposed to fully explore the prior knowledge to accelerate the new task learning. We also propose a novel dynamic graph weight router to reuse all previously learned knowledge to promote new task learning. The results demonstrate that the proposed approach achieves state-of-the-art performance.

\bibliographystyle{plain}
\bibliography{VAEGAN}



\section*{Supplementary material}

# Appendix for Incrementally Learning Multiple Diverse Data Domains Via Multi-Source Dynamic Expansion Model

November 22, 2024

# Contents

## A  The Additional Information for the Related Wrok

We introduce an innovative continual learning methodology that utilizes multiple backbones as the foundational backbone to enhance representation in multi-domain continual learning scenarios. In particular, we dynamically generate a new expert from the foundational backbone to tailor the general representation for new task acquisition. By amalgamating knowledge features from various domains and employing multi-head attention mechanisms alongside graph-layer architectures to prioritize and merge domain-specific features, our approach effectively synthesizes cross-task information. This technique minimizes task interference while alleviating the effects of catastrophic forgetting.

Furthermore, we employ a dynamic expert selection framework through a Router, which allows the model to adaptively choose the most pertinent experts according to the specific demands of the current task. The primary benefit of this strategy is its sparsity and selective activation: only the most relevant experts for each task are engaged, thereby minimizing computational overhead and memory usage. This approach mitigates the computational instability frequently associated with Mixture of Experts (MoE) models [4]. In contrast to MoE methods that depend on static expert selection to promote task information sharing, our method offers the necessary flexibility for optimal expert selection, preventing the engagement of irrelevant experts and consequently reducing the computational load.

When juxtaposed with the DER [1] and DER++ [1] methodologies, our model presents considerable benefits. For instance, the DER and DER++ typically utilize a memory buffer to mitigate forgetting in continual learning, with their efficacy often contingent upon the quality of the retained samples. Furthermore, these approaches are characterized by a singular and simplistic network architecture, which fails to deliver a robust representation for continual learning. In contrast, the proposed framework is capable of dynamically generating new experts to assimilate new tasks while preserving optimal performance on prior tasks by freezing all previously acquired parameters.

DAP [2]produces historical task data through instance-level prompts, effectively retains memories of prior tasks. However, it depends on the CLIP model and a replay

Table 1: Performance comparison from ablation studies: Tiny-Birds, Birds-Tiny, and Cifar100-Birds. For MSDEM, multi-head attention defaults to a single head. "In21k" indicates pretraining on ImageNet21K, while "ft-In1k" refers to fine-tuning on ImageNet1K "ViT_1 + ViT_2" denotes feature fusion through concatenation along the dim=1 axis of the output from two networks. The MSDEM$^2$ models use ViT_1 and ViT_2 as backbones. While the results reported in the paper for MSDEM$^2$ correspond to the configuration using ViT_1 and ViT_3. The notation 768*2-288 indicates an embedding input dimension of 768*2 and an output dimension of 288.

| Method | Tiny-Birds | | Birds-Tiny | | Cifar100-Birds | |
|---|---|---|---|---|---|---|
| | Avg. | Δ | Avg. | Δ | Avg. | Δ |
| In21k-ft-In1k (ViT_1) | 94.81 | -1.62 ↓ | 94.91 | -1.34 ↓ | 93.02 | -3.64 ↓ |
| In21k (ViT_2) | 83.78 | -12.65 ↓ | 81.51 | -14.74 ↓ | 80.51 | -16.15 ↓ |
| ViT-L-14 (ViT_3) | 87.43 | -9.00 ↓ | 85.86 | -10.39 ↓ | 86.98 | -9.68 ↓ |
| ViT_1 + ViT_2 | 92.55 | -3.88 ↓ | 93.08 | -3.17 ↓ | 89.71 | -6.96 ↓ |
| ViT_1 + ViT_2 + ViT_3 | 93.71 | -2.73 ↓ | 95.05 | -1.2 ↓ | 93.42 | -3.24 ↓ |
| MSDEM$^2$ (768*2-288) | **96.43** | **–** | **96.25** | **–** | **96.66** | **–** |
| –No Attention | 95.68 | -0.75 ↓ | 95.52 | -0.73 ↓ | 96.20 | -0.46 ↓ |
| –No Router | 95.94 | -0.49 ↓ | 96.19 | -0.06 ↓ | 96.34 | -0.32 ↓ |
| –No Embedding | 96.08 | -0.36 ↓ | 96.08 | -0.17 ↓ | 96.19 | -0.47 ↓ |
| MSDEM$^3$ (768*3-768) | **96.71** | **–** | **96.66** | **–** | **97.11** | **–** |
| –No Attention | 96.73 | **+0.02 ↑** | 96.68 | -0.43 ↓ | 97.05 | -0.06 ↓ |
| –No Router | 96.64 | -0.07 ↓ | 96.43 | -0.23 ↓ | 96.97 | -0.14 ↓ |
| –No Embedding | 97.05 | **+0.34 ↑** | 97.01 | **+0.35 ↑** | 97.25 | **+0.14 ↑** |

mechanism, leading to increased computational demands. Additionally, the prompts generated may be limited by the constraints of the original pre-trained model, which can hinder the adaptability of task transfer across various domains. In contrast, RanPAC mitigates computational complexity via random projection, integrating pre-trained models for task learning. Nevertheless, it relies on the buffering of previous task data and employs generative replay to avert catastrophic forgetting, which incurs storage overhead and lacks the dynamic expert selection capability that our approach provides.

StarPrompt [3] implements task transfer learning via prompt tuning. For each new task, the model fine-tunes the prompt template to enhance alignment with the specific task, thereby minimizing extensive alterations to the network architecture. Although this approach effectively utilizes pre-trained models and circumvents significant changes to the network structure, the optimization of the prompt template during training substantially escalates computational demands. As the number of tasks proliferates, the model is required to continuously refine prompts and create new task-specific templates, leading to an accumulation of computational load. This issue is

Table 2: Performance comparison from ablation studies: Birds-Cifar100, TinyImageNet-Cifar100-Birds, and TinyImageNet-Cifar100-Birds-Cifar10.

| Method | Birds-Cifar100 | | Tiny-Cifar100-Birds | | Tiny-C100-B-C10 | |
|---|---|---|---|---|---|---|
| | Avg. | Δ | Avg. | Δ | Avg. | Δ |
| In21k-ft-In1k (ViT_1) | 94.04 | -2.64 ↓ | 92.66 | -2.73 ↓ | 93.37 | -1.95 ↓ |
| In21k (ViT_2) | 79.80 | -16.88 ↓ | 79.99 | -15.4 ↓ | 81.98 | -13.34 ↓ |
| ViT-L-14 (ViT_3) | 85.36 | -11.32 ↓ | 87.59 | -7.8 ↓ | 87.51 | -7.82 ↓ |
| ViT_1 + ViT_2 | 91.19 | -5.49 ↓ | 90.64 | -4.75 ↓ | 90.95 | -4.37 ↓ |
| ViT_1 + ViT_2 + ViT_3 | 93.11 | -3.57 ↓ | 92.73 | -2.66 ↓ | 92.68 | -2.64 ↓ |
| **MSDEM$^2$ (768*2-288)** | **96.68** | – | **95.39** | – | **95.32** | – |
| –No Attention | 96.13 | -0.55 ↓ | 94.28 | -1.11 ↓ | 94.51 | -0.81 ↓ |
| –No Router | 96.22 | -0.46 ↓ | 94.94 | -0.45 ↓ | 94.78 | -0.54 ↓ |
| –No Embedding | 96.25 | -0.43 ↓ | 94.8 | -0.59 ↓ | 94.71 | -0.61 ↓ |
| **MSDEM$^3$ (768*3-768)** | **97.02** | – | **95.93** | – | **95.85** | – |
| –No Attention | 96.95 | -0.07 ↓ | 95.77 | -0.15 ↓ | 95.80 | -0.05 ↓ |
| –No Router | 96.86 | -0.16 ↓ | 95.83 | -0.10 ↓ | 95.74 | -0.11 ↓ |
| –No Embedding | 97.27 | **+0.25 ↑** | 96.18 | **+0.25 ↑** | 96.30 | **+0.45 ↑** |

Table 3: Performance comparison across Tiny-Birds, Birds-Tiny, Cifar10-Birds, and Birds-Cifar10 under different combinations of backbone configurations.

| Method | Tiny-Birds | | Birds-Tiny | | Cifar10-Birds | | Birds-Cifar10 | |
|---|---|---|---|---|---|---|---|---|
| | Avg. | Δ | Avg. | Δ | Avg. | Δ | Avg. | Δ |
| MSDEM$^2$ (ViT_1+ViT_2) | 95.01 | 2.82 ↓ | 96.25 | -1.49 ↓ | 97.73 | -1.59 ↓ | 97.89 | -1.55 ↓ |
| **MSDEM$^2$ (ViT_1+ViT_3)** | **97.83** | – | **99.74** | – | **99.32** | – | **99.44** | – |
| + ViT_In21k | 96.71 | -1.12 ↓ | 96.72 | -1.02 ↓ | 99.11 | -0.21 ↓ | 99.13 | -0.31 ↓ |
| + ViT-B-16 | 97.64 | -0.19 ↓ | 97.69 | -0.05 ↓ | 99.44 | **0.12 ↑** | 99.64 | **0.21 ↑** |
| + ViT-B-32 | 97.79 | -0.04 ↓ | 97.73 | -0.01 ↓ | 99.26 | -0.06 ↓ | 99.34 | -0.10 ↓ |
| + ResNet-50 | 97.65 | -0.18 ↓ | 97.63 | -0.11 ↓ | 99.35 | **0.03 ↑** | 99.35 | 0 |
| + ResNet-101 | 97.61 | -0.22 ↓ | 97.58 | -0.16 ↓ | 99.52 | **0.2 ↑** | 99.35 | -0.09 ↓ |

particularly exacerbated when dealing with high-dimensional data, where computational efficiency is adversely affected. Additionally, prompt optimization necessitates further computations on the original pre-trained model, which is frequently large and resource-intensive. For instance, the CLIP model is inherently computationally intensive, rendering prompt optimization a potential bottleneck in large-scale, multitask learning environments.

In contrast, our model significantly minimizes computational overhead by utilizing dual backbones and dynamic expert selection. The multi-backbone architecture concurrently captures both global and local features, thereby enhancing the efficiency of feature learning and circumventing the repetitive prompt adjustments typical of

Table 4: Performance comparison across Cifar100-Birds, Birds-Cifar100, T-C10-B, T-C100-B, and T-C100-B-C10 under different combinations of backbone configurations.

| Method | C100-Birds | | Birds-C100 | | T-C10-B | | T-C100-B | | T-C100-B-C10 | |
|---|---|---|---|---|---|---|---|---|---|---|
| | Avg. | Δ | Avg. | Δ | Avg. | Δ | Avg. | Δ | Avg. | Δ |
| MSDEM$^2$ (ViT_1+ViT_2) | 95.93 | 2.1↓ | 96.01 | 2.13↓ | 95.12 | 2.62↓ | 94.05 | 2.87↓ | 94.31 | 2.78↓ |
| MSDEM$^2$ (ViT_1+ViT_3) | **98.03** | – | **98.14** | – | **97.74** | – | **96.92** | – | **97.09** | – |
| + ViT_In21k | 97.11 | 0.92↓ | 97.02 | 1.12↓ | 97.15 | 0.59↓ | 96.15 | 0.77↓ | 95.92 | 1.17↓ |
| + ViT-B-16 | 98.00 | 0.03↓ | 98.06 | 0.08↓ | 97.83 | **0.09↑** | 97.07 | **0.15↑** | 96.98 | 0.11↓ |
| + ViT-B-32 | 97.84 | 0.19↓ | 98.17 | **0.03↑** | 97.59 | 0.15↓ | 97.11 | **0.19↑** | 97.01 | 0.08↓ |
| + ResNet-50 | 97.99 | 0.01↓ | 97.93 | 0.21↓ | 97.68 | 0.06↓ | 96.90 | 0.02↓ | 97.03 | 0.06↓ |
| + ResNet-101 | 98.03 | 0 | 97.83 | 0.31↓ | 97.66 | 0.08↓ | 96.89 | 0.03↓ | 96.93 | 0.16↓ |

Table 5: Basic information and transformations for datasets.

| Dataset | Original Size | Phase | Resizing & Cropping |
|---|---|---|---|
| TinyImageNet | $3 \times 64 \times 64$ | Training | Resize 300 → 224, Random crop |
| | | Testing | Resize 256 → 224, Center crop |
| CIFAR100 | $3 \times 32 \times 32$ | Training | Random resized crop 224 |
| | | Testing | Resize 224 |
| Birds-200 | $3 \times 224 \times 224$ | Training | Random crop (padding 4) |
| | | Testing | No resizing or cropping |
| CIFAR10 | $3 \times 32 \times 32$ | Training | Resize 224, Random crop (padding 28) |
| | | Testing | Resize 224 |

StarPrompt. Furthermore, the dynamic expert selection, executed through the Gumbel-Softmax mechanism, engages only the most pertinent experts for the specific task at hand, thus eliminating unnecessary computations and markedly reducing resource consumption. This leads to considerable improvements in computational efficiency, especially in large-scale data and multitask learning contexts. Consequently, when compared to the prompt optimization strategy employed by StarPrompt, our model presents a clear advantage in terms of computational efficiency.

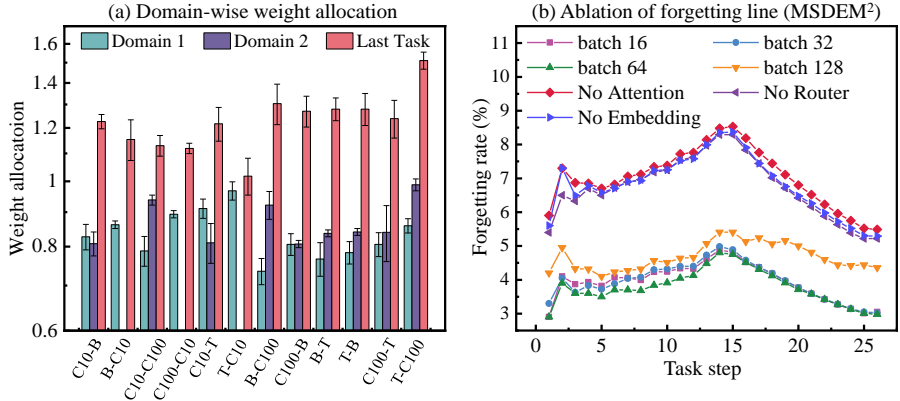

Figure 1: (a) Model weight allocation across historical domains, the current domain, and the current task for various permutation schemes; (b) Forgetting curves from ablation studies and the influence of different batch sizes on model performance. MSDEM$^2$ in (c)-(f) denotes using ViT_1 and ViT_3 as backbones.

# B  The Additional Information for the Experiment Settings

## B.1  Training details

We introduce a novel multi-backbone network architecture tailored for multi-task learning, integrating a dynamic feature selection mechanism to enhance performance. The proposed architecture comprises three pretrained Vision Transformer (ViT) backbones. The first backbone (ViT_1) was pre-trained on the ImageNet-21K dataset and fine-tuned on ImageNet-1K. The second backbone (ViT_2) was trained exclusively on ImageNet-21K. The third backbone is a variant of the ViT-L/14 model from OpenAI's CLIP, pre-trained on a large-scale image-text paired dataset. To preserve the prior knowledge encapsulated in these backbones, their parameters remain frozen during training, serving exclusively as feature extractors. Each backbone outputs a 768-dimensional feature vector, concatenated to produce a 1536-dimensional vector (dual backbones) or a 2304-dimensional vector (triple backbones).

To enable dynamic feature allocation across tasks, a routing module based on the

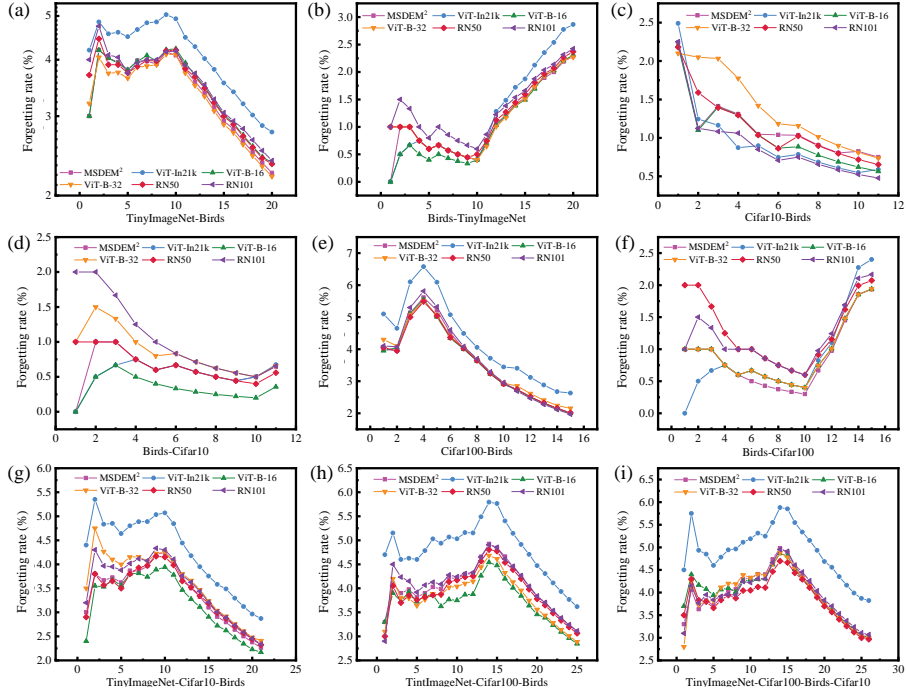

Figure 2: (a)-(i) Performance comparison of forgetting curves for MSDEM$^2$ and its extended version with a third backbone. The backbone for MSDEM$^2$ consists of ViT_1 and ViT_3, while we compare three ViT-type backbones with two ResNet-type backbones. "RN" refers to "ResNet". The data results include six types of dual-domain task scenarios, two types of three-domain task scenarios, and one type of four-domain task scenario.

Noisy Top-K Softmax mechanism was introduced. This module assigns scores to features and utilizes a Gumbel-Softmax-based selection mechanism to sparsely identify the most relevant Top-K task-specific features. Moreover, each task is equipped with a lightweight attention module and a graph module to refine task-specific feature representations. The attention module reduces feature dimensionality from 1536 or 2304 to 288, effectively lowering computational overhead while improving feature extraction efficiency.

For classification tasks, each task is assigned an independent linear classification head, optimized using the Adam optimizer, dynamically adjusted via a cosine annealing scheduler. The routing, attention, and graph modules are optimized using distinct

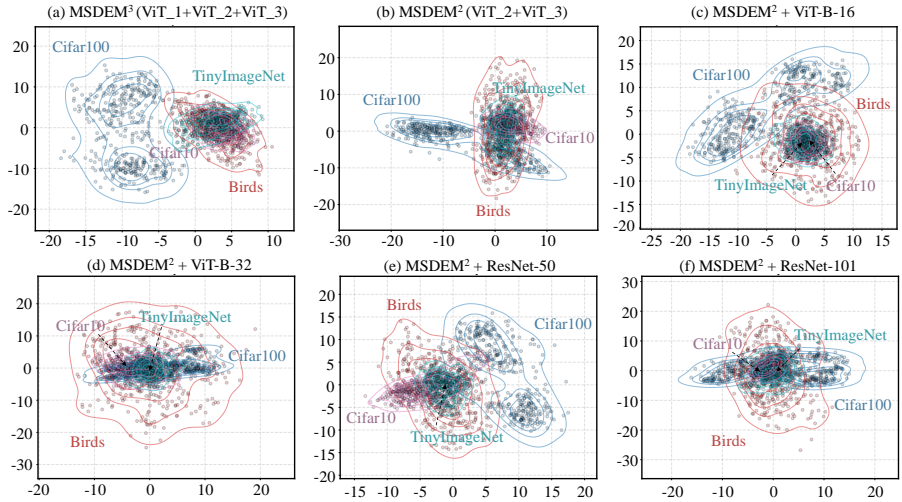

Figure 3: (a)-(f) PCA feature distribution of experts across four task domains under different backbone combinations. MSDEM$^2$ in (c)-(f) denotes using ViT_1 and ViT_3 as backbones.

Adam optimizers and cosine annealing schedulers. To improve the robustness of feature selection, the routing module is coupled with a Gaussian noise mechanism.

## B.2 Hyperparameter selection

In our model, we employ distinct Adam optimizers for the DEAM, Router, Graph Block, and Classifier components, with learning rates selected from the range $\{10^{-4}, 10^{-2}\}$. To mitigate the risk of excessive parameter growth, the output dimension of the Embedding layer in DEAM is set to 288. The number of heads in all multi-head attention mechanisms is uniform, and the search space for this hyperparameter spans $\{4, 8, 16, 32\}$.

For hyperparameter optimization, we adopt Bayesian optimization, leveraging a distributed approach and using the validation set for performance evaluation. In contrast to random search and grid search, Bayesian optimization applies Bayesian statistical principles, iteratively refining hyperparameter selection by incorporating insights from prior search results. This methodology not only improves search efficiency but also facilitates more effective convergence towards the optimal solution.

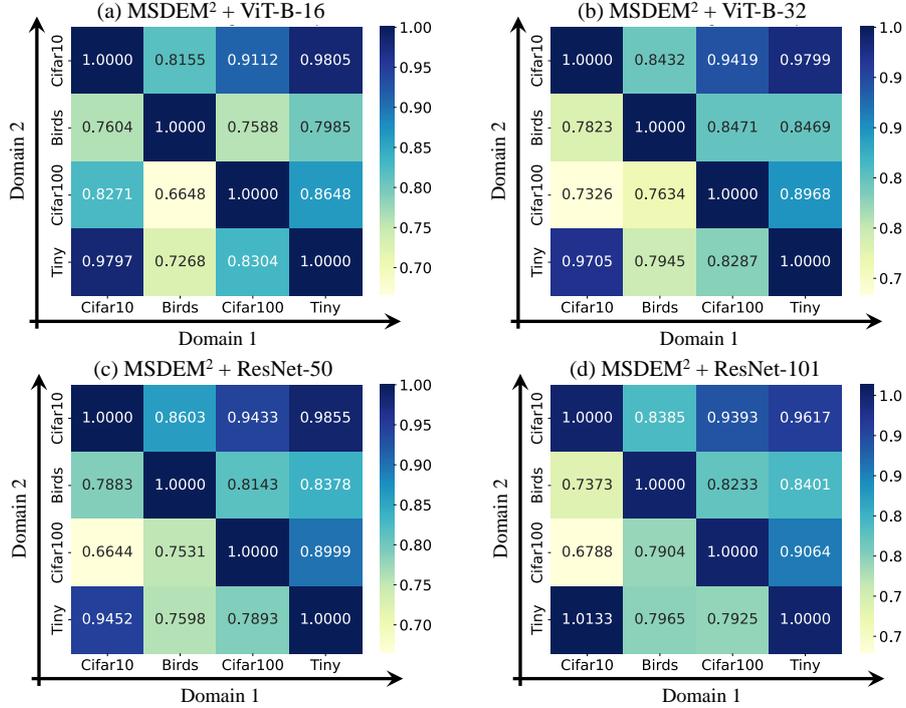

Figure 4: (a)-(d) Weight distribution across 16 dual-domain task scenarios under different backbone combination schemes, with the weight dependency of the domain itself set to 1. MSDEM$^2$ in (a)-(d) denotes using ViT_1 and ViT_3 as backbones.

## B.3 Dataset configurations

The experimental setup encompasses a multi-domain, multi-task sequence spanning four distinct domains: TinyImageNet, CIFAR100, Birds-200, and CIFAR10. In T-C100-B-C10 task configuration, TinyImageNet is assigned to the first 10 tasks (tasks 0–9), CIFAR100 to tasks 10–14, Birds to tasks 15–24, and CIFAR10 to the final task (task 25). All datasets were resized to a standardized resolution of 224×224. We summarize the data augmentation details in Table.5. During training, data augmentation techniques such as random cropping, horizontal flipping, and Gaussian blur were employed, while testing utilized centre cropping for standardization.

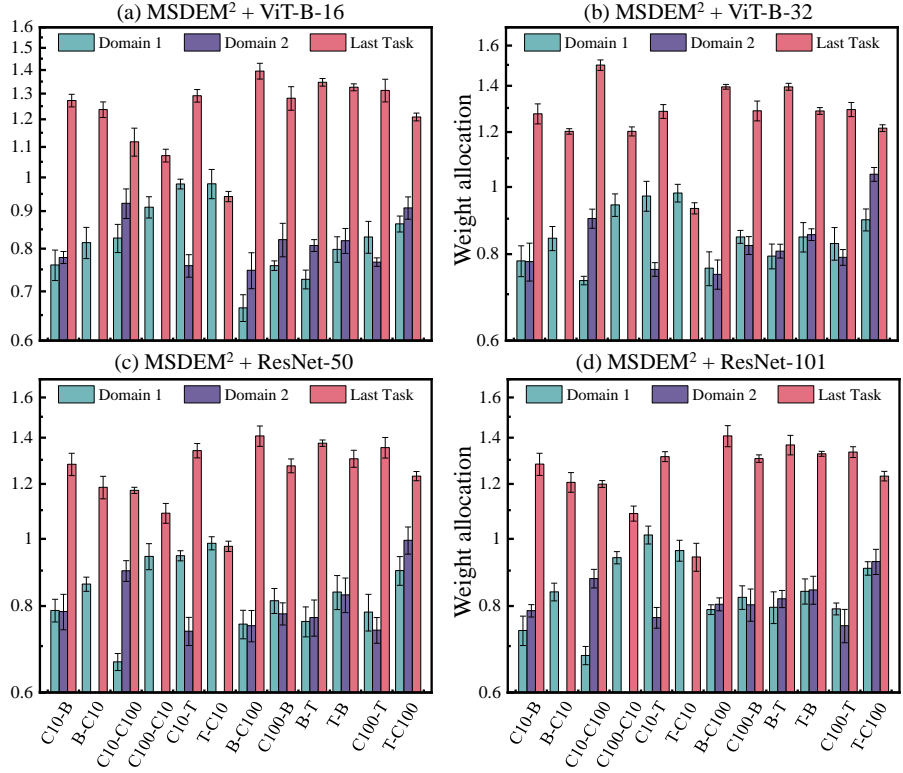

Figure 5: (a)-(d) Weight allocation of different domains across 16 dual-domain task scenarios under different backbone combination schemes. In all scenarios, the weight of the last task is the highest, and the relative weight between domain1 and domain2 is dependent on their task order.

### B.4 Device configurations

All experiments were conducted on the same hardware environment running Ubuntu 22.04.2 LTS, with 32 GB of RAM, and Intel(R) Xeon(R) CPU E5-2683 v4 processor @ 2.10GHz. A single NVIDIA GeForce RTX4090 GPU provides the computing acceleration in experiments.

All tasks were trained with a uniform batch size of 32, and a fixed random seed of 1993 was used to ensure experimental reproducibility. The cross-entropy loss function was employed, and average accuracy served as the primary evaluation metric. Experimental results demonstrate that the proposed multi-backbone architecture, in conjunc-

tion with dynamic feature selection mechanisms, significantly improves multi-domain learning performance while effectively mitigating catastrophic forgetting.

## C  The Additional Ablation Studies

In the following sections, we provide additional ablation studies to evaluate the performance of the proposed framework under different configurations.

### C.1  Analysis of Backbone and Attention

We introduce an attention mechanism to fuse the output features of multiple ViTs, thereby integrating the knowledge learned by pre-trained models across different domains. Specifically, we consider creating several baselines, where each only uses a unique ViT backbone. We provide the analysis results in Table.1 and Table.2, where MSDEM[2] and MSDEM[3] denotes the proposed approach with dual-ViT and triple-ViT models, respectively. The results demonstrate that neither a single ViT nor a direct summation of features can effectively enhance model performance in all task configurations. In contrast, the attention mechanism as a fusion strategy can achieve better performance, demonstrating its effectiveness.

### C.2  Analysis of Embedding

The number of training parameters required by our proposed attention mechanism is highly dependent on the dimensionality of the input features, with a computational complexity of $O(n^2)$. To avoid an explosion in training parameters and to manage redundant features, we introduce an embedding layer as a dimensionality reduction method. In our experimental setup, each ViT's output feature dimension is set to 768. For the dual-ViT model architecture, the embedding layer reduces the feature dimension to 288, while for the triple-ViT architecture, it remains at 768. Table. 1 and Table. 2 compare model performance with and without dimensionality reduction through the embedding layer. In the dual-ViT architecture, nearly all results show performance degradation without embedding, whereas the trend is reversed in the triple-ViT archi-

tecture. Nevertheless, omitting dimensionality reduction in the triple-ViT setup results in an exponential increase in training parameters, which does not yield a reasonable cost-benefit ratio relative to the performance gains achieved.

Figure. 1.(a) further summarizes the weights from all permutation schemes, indicating that the model consistently prioritizes knowledge from the currently trained domain, maximizing weight allocation for the current task, while weights for historical domains follow the aforementioned distribution pattern. Figure. 1.(b) displays forgetting curves that compare the effects of various components in the ablation study, as well as the impact of different batch sizes on model performance. These results highlight the model's robustness to batch size variations and demonstrate that the Router enhances performance by modulating the dependency on prior domain knowledge.

### C.3 Analysis of Backbone selection

In the ablation study, we observed that ViT_2 exhibited the lowest performance among all the ViT variants. To further explore the impact of different ViT combinations on model performance, we conducted four additional experiments, where ViT_2 was replaced by pre-trained ViT-B-16, ViT-B-32, ResNet-50, and ResNet-101. The results of these experiments are summarized in Table 3 and 4, and the forgetting curves for the different backbone configurations are presented in Figure. 2. The relevant URLs for pre-trained models are listed at the end of the appendix.

Building upon these combinations, we analyzed the feature output distributions of the experts and visualized the results. The corresponding feature distributions and weight heatmaps are provided in Figure. 3 and Figure. 4, respectively. Overall, incorporating new high-performance backbones enhanced the model's overall performance, with some configurations surpassing the original MSDEM$^2$. However, the independence and compactness of the feature distributions were slightly diminished compared to MSDEM$^2$. Additionally, we visualized the weight allocation for the two domains under various task settings in Figure 5. The model architecture with three backbones, shown in Figure 5, maintained consistent weight distribution patterns across tasks.

- **RN50**: https://openaipublic.azureedge.net/clip/models/afeb0e10f9e5a86da6080e35

RN50.pt

- **RN101**: `https://openaipublic.azureedge.net/clip/models/`
  `8fa8567bab74a42d41c5915025a8e4538c3bdbe8804a470a72f30b0d94fab599/`
  `RN101.pt`

- **ViT-B/32**: `https://openaipublic.azureedge.net/clip/models/`
  `40d365715913c9da98579312b702a82c18be219cc2a73407c4526f58eba950af/`
  `ViT-B-32.pt`

- **ViT-B/16**: `https://openaipublic.azureedge.net/clip/models/`
  `5806e77cd80f8b59890b7e101eabd078d9fb84e6937f9e85e4ecb61988df416f/`
  `ViT-B-16.pt`

- **ViT-L/14**: `https://openaipublic.azureedge.net/clip/models/`
  `b8cca3fd41ae0c99ba7e8951adf17d267cdb84cd88be6f7c2e0eca1737a03836/`
  `ViT-L-14.pt`



\end{document}